\documentclass[journal]{IEEEtran}
\usepackage{cite}  % optional but often used with IEEE
\usepackage[utf8]{inputenc}

\usepackage{graphicx}
\usepackage{amsmath}
\usepackage{amssymb}
\usepackage{booktabs}
\usepackage{multirow}
\usepackage{multicol}
\usepackage{float}
\usepackage{todonotes}
\usepackage{comment}
\usepackage{tabularx}
\usepackage{subcaption}
\usepackage{booktabs}
\usepackage[breaklinks,colorlinks, citecolor=blue]{hyperref}
\usepackage{tikz}
\usepackage[capitalize]{cleveref}
\crefname{section}{Sec.}{Secs.}
\Crefname{section}{Section}{Sections}
\Crefname{table}{Table}{Tables}
\crefname{table}{Tab.}{Tabs.}
\RequirePackage{xspace}
\makeatletter
\DeclareRobustCommand\onedot{\futurelet\@let@token\@onedot}
\def\@onedot{\ifx\@let@token.\else.\null\fi\xspace}

\makeatother

\begin{document}

%\title{Deep Clustering with Soft Constraints: Leveraging Spatial and Depth Constraints for Martian Terrain Recognition}
\title{Enhancing Martian Terrain Recognition with Deep Constrained Clustering}

\author{Tejas~Panambur,~\IEEEmembership{Member,~IEEE,}
        Mario~Parente,~\IEEEmembership{Senior Member,~IEEE}
        % <-this % stops a space
\thanks{Tejas Panambur is with the Department
of Electrical and Computer Engineering, University of Massachusetts, Amherst,
MA, 01003 USA e-mail: (tpanambur@umass.edu).}% <-this % stops a space
\thanks{}% <-this % stops a space
\thanks{}}

\markboth{IEEE TRANSACTIONS ON GEOSCIENCE AND REMOTE SENSING}%
{Panambur \MakeLowercase{\textit{et al.}}: }
\maketitle

\begin{abstract}
%The Mars Science Laboratory (MSL) Curiosity and Perseverance rovers have played a crucial role in capturing images of the Martian terrain, which has been instrumental in studying various aspects such as topography, geomorphology, Martian paleoclimate, and habitability. 
Martian terrain recognition is pivotal for advancing our understanding of topography, geomorphology, paleoclimate, and habitability.  
While deep clustering methods have shown promise in learning semantically homogeneous feature embeddings from Martian rover imagery, the natural variations in intensity, scale, and rotation pose significant challenges for accurate terrain classification. To address these limitations, we propose Deep Constrained Clustering with Metric Learning (DCCML), a novel algorithm that leverages multiple constraint types to guide the clustering process. DCCML incorporates soft must-link constraints derived from spatial and depth similarities between neighboring patches, alongside hard constraints from stereo camera pairs and temporally adjacent images.  Experimental evaluation on the Curiosity rover dataset (using $150$ clusters) demonstrates that DCCML increases homogeneous clusters by $16.7\%$ while reducing the Davies–Bouldin Index from $3.86$ to $1.82$ and boosting retrieval accuracy from $86.71\%$ to $89.86\%$. This improvement enables more precise classification of Martian geological features, advancing our capacity to analyze and understand the planet's landscape.

\end{abstract}


\IEEEpeerreviewmaketitle

\section{Introduction}

\IEEEPARstart{U}{nderstanding} the Martian landscape is a primary objective of Mars surface science, pursued through various robotic missions, including NASA's Mars Science Laboratory (Curiosity) \cite{msl} and Mars-2020 (Perseverance) \cite{mars2020}. These rovers are equipped with advanced imaging systems, such as Mastcam and Mastcam-Z, which capture high-resolution images of the diverse terrains around their landing sites. Leveraging these imaging capabilities, accurate terrain recognition plays a pivotal role in advancing the science goals of these missions. It enables the rovers to navigate safely and efficiently across the Martian surface, avoiding hazards such as large rocks, steep slopes, and soft sand that could impede their progress or cause damage. Additionally, terrain recognition aids in selecting optimal sites for conducting experiments and collecting samples, ensuring the success of these groundbreaking missions. 

Most importantly, by accurately identifying and analyzing different types of terrain, scientists can reveal information about rock formation processes, past environmental conditions, and potential habitats that may have supported ancient microbial life (such as ancient riverbeds or paleo-lakes), enhancing their ability to help understand the planet's geological history \cite{msl, Bell2017-ia, Malin2017-vg}.

%One-two paragraphs to explain Existing methods 
Extensive efforts have been made to apply supervised deep learning approaches to terrain classification tasks \cite{scoti, chakravarthy2021mrscatt, mslv1, mslv2, wang2021semi, gonzalez2018deepterramechanics, PALAFOX201748, spoc, ai4mars}. These methods typically classify terrains based on visually discernible features such as texture, color, and morphology. While effective for operational objectives like navigation and obstacle avoidance, they lack the granularity necessary for advancing scientific investigations. A key limitation lies in their reliance on annotations from non-experts, which constrains classifications to broad morphological categories (e.g., soil, sand, bedrock, float rock) that fail to capture the nuanced geological variations critical for understanding planetary processes. Additionally, the vast volume of high-resolution terrain data generated by Mars rovers far exceeds the capacity of the limited number of planetary geologists available to annotate, analyze, and classify it manually. This scarcity of experts, combined with the inefficiencies and inconsistencies of manual methods, underscores the urgent need for automation to address these challenges.

To address these challenges, limited approaches have been developed, primarily focusing on self-supervised representation learning methods such as contrastive learning \cite{jplcontrastive}, or our prior work on deep clustering for deriving feature embeddings from unlabeled data \cite{tejas1, tejas2, tejasearth}. The contrastive training process described in \cite{jplcontrastive} groups random augmented views of the same source image or cluster and evaluates classification performance on a few coarse classes by attaching a linear layer to the learned representations. While effective for certain general-purpose tasks, these methods, like their supervised counterparts, suffer from the same limitations due to their reliance on labeled data constrained to broad perceptual categories. As a result, they fail to capture fine-grained geological variations essential for scientific analysis, making them unsuitable for planetary exploration. 

In contrast, deep clustering methods provide a label-free alternative for learning structured feature embeddings that capture fine-grained geological patterns from unlabeled data. Our previous work has demonstrated that deep clustering can iteratively refine feature embeddings by alternately clustering Martian terrain images and optimizing the network using pseudo-labels generated from the clustering process \cite{tejas1, tejas2, tejasearth}. These methods enable the discovery of meaningful terrain groups without reliance on predefined class labels, facilitating a more detailed and scientifically relevant characterization of Martian geology. As a result, \cite{tejasearth} provides a comprehensive taxonomy, developed in collaboration with a planetary geologist for categorizing Martian terrain observed in Curiosity Mastcam images. A detailed discussion of our approach is presented in \cref{ss:relatework}.

Despite these advancements, clustering performance remains constrained by the significant natural variations present on the Martian surface, including differences in intensity, scale, rotation, illumination, and other intrinsic factors within similar terrain patches. These variations can cause geologically similar patches to be projected farther apart in the feature space, while patches with superficial visual resemblance may be incorrectly grouped together, leading to inconsistent clustering results. 

\begin{figure*}[htb]
\centering
 \begin{minipage}{.48\textwidth}
    \includegraphics[width=\textwidth, height=3cm]{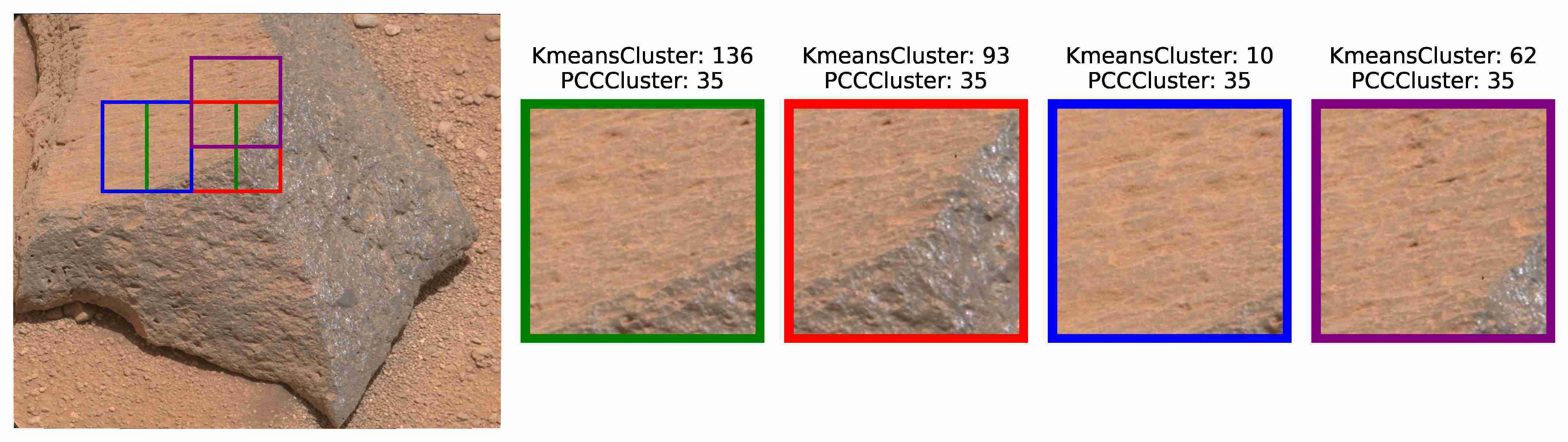}
    \subcaption{}
\end{minipage}
\begin{minipage}{.48\textwidth}
    \includegraphics[width=\textwidth, height=3cm]{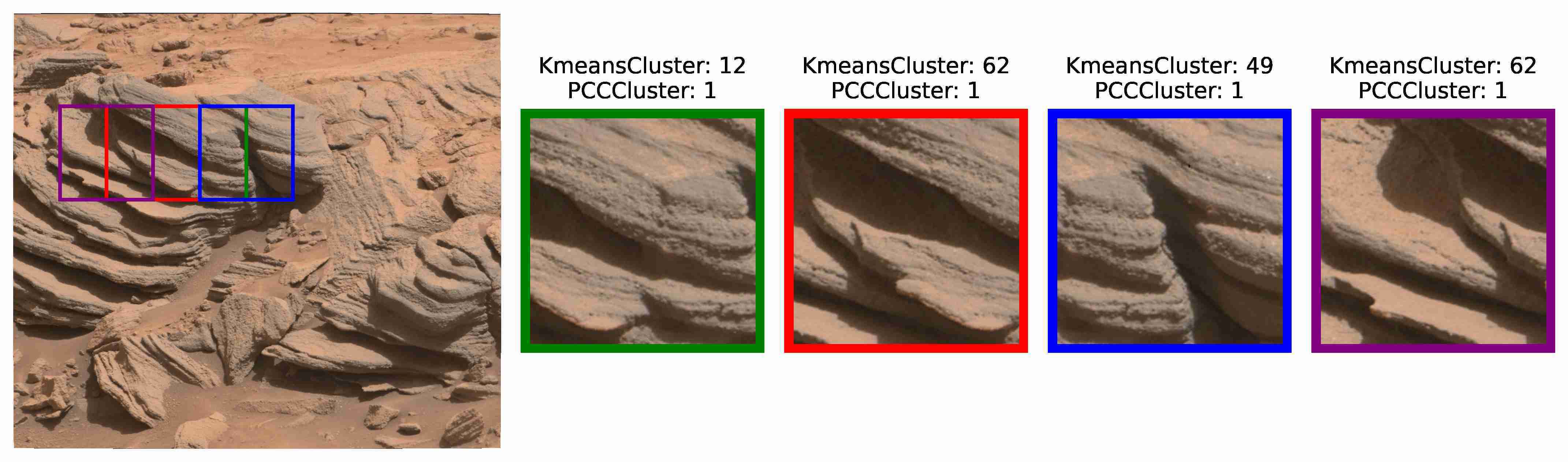}
    \subcaption{}
\end{minipage}
\centering
  \begin{minipage}{.48\textwidth}
    \includegraphics[width=\textwidth, height=3cm]{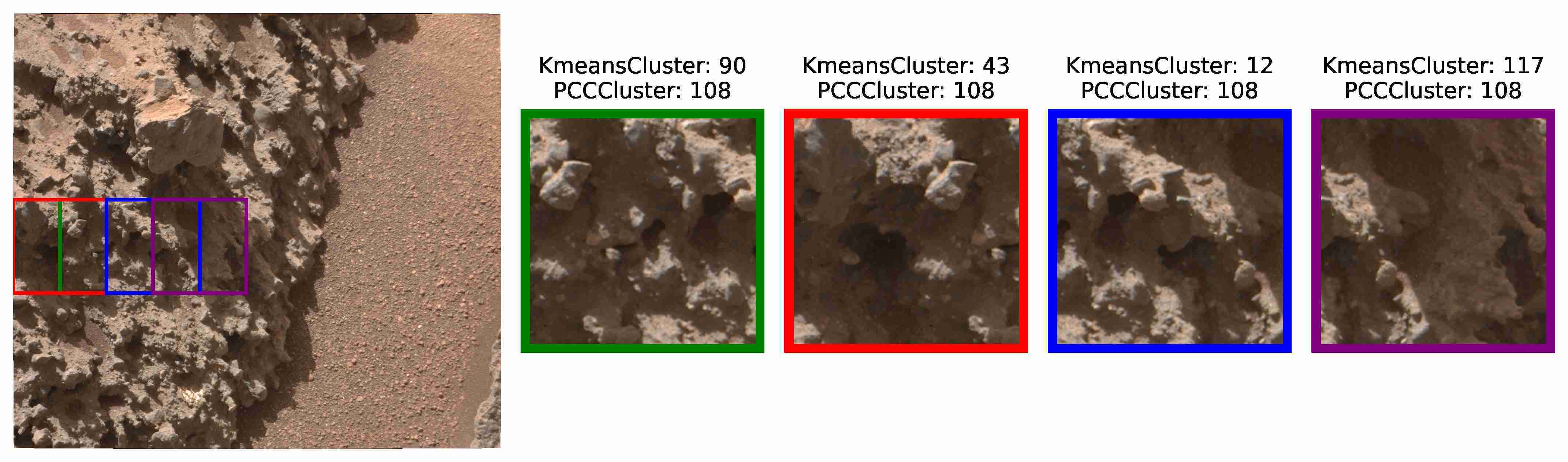}
    \subcaption{}
\end{minipage}
\begin{minipage}{.48\textwidth}
    \includegraphics[width=\textwidth, height=3cm]{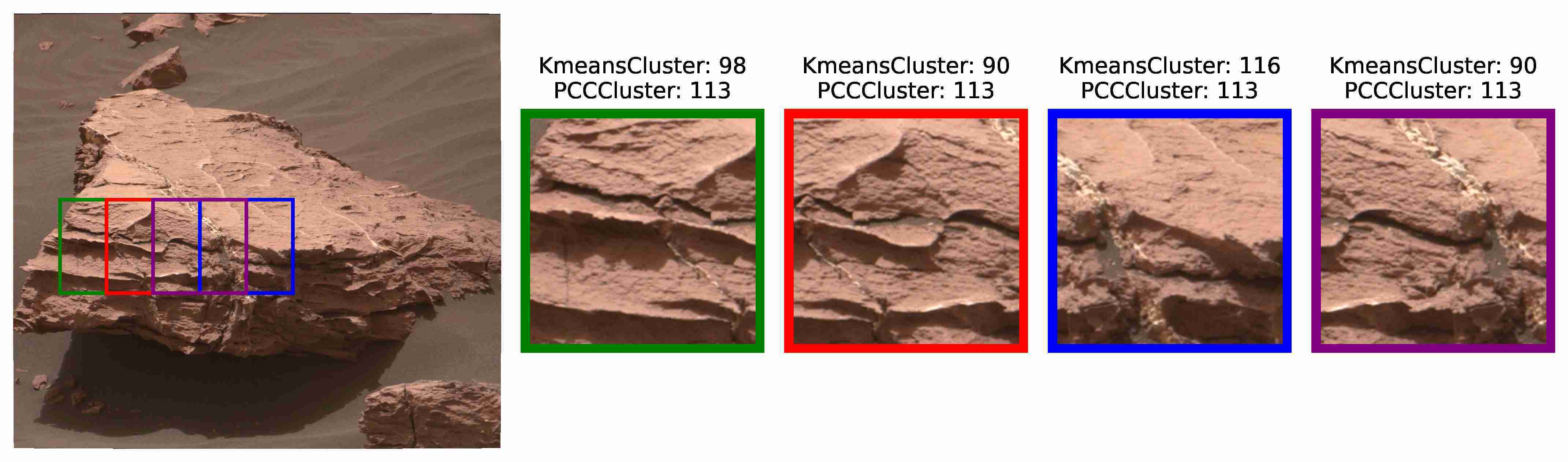}
    \subcaption{}
\end{minipage}
 \caption{The figure shows 4 Martian Terrain images along with 4 patches extracted from each of these images. Displaying a Martian terrain image alongside similar terrain patches from the same image, showcasing diverse textures within the same terrain class.}
 \label{fig:NebrSim}
\end{figure*}

For example, \cref{fig:NebrSim} presents four images of distinct Martian terrains, with adjacent patches highlighted in different colors and enlarged to the right of each original image. In \cref{fig:NebrSim}.a, variations in dust coverage affect neighboring patches, while \cref{fig:NebrSim}.b reveals differences in illumination, shadowing, and layering textures. Similarly, \cref{fig:NebrSim}.c and \cref{fig:NebrSim}.d highlight variations in shadow intensity, nodularity, and vein thickness. Despite originating from the same rock formation, our previous clustering methods incorrectly partitioned these patches into separate clusters due to these visual inconsistencies. 

Furthermore, the left subplot of \cref{fig:cusres1} and \cref{fig:cusres2} demonstrates how clustering results are influenced by these variations, often leading to the formation of clusters based on superficial visual similarities rather than intrinsic geological properties. For example, in \cref{fig:cusres1}.a, clustering is predominantly driven by illumination conditions, whereas in \cref{fig:cusres2}.a, similarity in hue dictates the grouping. These artifacts highlight the limitations of unsupervised clustering, where variations such as lighting, dust coverage, and color differences mislead the learning process. 

%To ensure scientifically meaningful clustering, methods must align with the validated planetary science taxonomy \cite{tejasearth}, ensuring that learned representations capture geologically significant features rather than irrelevant visual artifacts, thereby improving classification accuracy and scientific utility. 

To mitigate the effects of superficial variations, contrastive learning approaches \cite{jplcontrastive} employ data augmentations to learn representations that are invariant to changes in illumination, scale, rotation, and other external factors. The objective of these augmentations is to ensure that models prioritize underlying geological features rather than being influenced by irrelevant visual artifacts. To achieve this, transformations such as cropping, scaling, and color adjustments are applied to enhance the robustness of learned representations. However, determining appropriate augmentations without labeled data is inherently challenging. In the absence of labels, augmentation strategies can introduce spurious transformations that distort geological features rather than improving model generalization. For instance, scaling a pebble may cause it to resemble a floating rock, while scaling a laminated rock could shift its categorization from strongly laminated to weakly laminated. Moreover,  Martian terrain is inherently complex, exhibiting natural variations in dust coverage, embedded pebbles, and fluctuating shadow conditions, making it difficult to design augmentations that preserve geological integrity without resorting to more sophisticated deep learning-based image generation techniques.

In contrast, our approach aims to overcome these limitations by directly learning from intrinsic variations in Martian terrain rather than relying on augmentations that risk distorting geological features. This paper proposes an improved deep-constrained clustering algorithm (DCCML) that integrates multiple constraints to guide the clustering process. Unlike contrastive learning, which enforces invariance through augmentations, our approach ensures that clusters align with true geological formations by leveraging spatial, depth, stereo, and RSM constraints.

To achieve this, our approach automatically generates pairwise constraints based on spatial proximity and depth similarity within an image, as detailed in \cref{subsec:genc}. The spatial constraints involve selecting patches that are spatially proximate to each other. Despite their spatial proximity, the Mastcam image exhibits patches with terrain at various depths, as observed in \cref{fig:NebrSim}. Therefore, we impose spatial proximity and constrain the patches to be within a similar depth range. While these constraints improve clustering accuracy, certain regions present additional challenges where terrain transitions are gradual rather than discrete. In such cases, individual patches may contain a blend of multiple terrain types. To mitigate this issue, we eliminate mixed patches that potentially contain multiple terrain types, as discussed in \cref{subsec:preprocessing}.

Further, the Mastcam stereo imaging system introduces additional challenges due to differences in resolution and perspective. The 100mm right-eye camera provides high-resolution, narrow field-of-view imaging, while the 34mm left-eye camera captures wider field-of-view images at a lower resolution. These inherent resolution and perspective differences caused our previous approach to misclassify corresponding left and right image pairs into separate clusters approximately $80\%$ of the time, likely due to scale mismatches and blurring effects. To address this, we identify corresponding patch regions between the right and left-eye images depicting the same Martian target and establish constraints to ensure their consistent clustering. Additionally, we introduce RSM constraints, which establish relationships between sequentially captured images that depict the same target area from slightly different viewpoints or under varying illumination conditions. 
%These constraints improve clustering stability, ensuring that geologically similar regions remain consistently grouped. 

By integrating these constraints, DCCML ensures that clustering is driven by geological consistency rather than superficial visual similarities, aligning the learned representations with scientifically validated planetary taxonomies and improving both classification accuracy and geologic interpretability.

\begin{comment}
\begin{figure*}[t]
    \centering
    \begin{tabular*}{\textwidth}
    {@{\extracolsep{\fill}}cc@{}}
        \centering
        \begin{tabular}
        {p{0.48\linewidth}p{0.48\linewidth}}
            \includegraphics[width=1\linewidth]{NebrSim/1099MR0048670020600851E01_DRCL_with_borders.jpg} &
            \includegraphics[width=1\linewidth]{NebrSim/0349MR0014180000301163E01_DRCL_with_borders.jpg} &
            \includegraphics[width=1\linewidth]{NebrSim/1407MR0068890200702114C00_DRCL_with_borders.jpg} &
            \includegraphics[width=1\linewidth]{NebrSim/1521MR0077460040204732E01_DRCL_with_borders.jpg} &
        \end{tabular}
      
    \end{tabular*}
    \caption{The figure shows 4 Martian Terrain images along with 4 patches extracted from each of this image. Displaying a Martian terrain image alongside similar terrain patches from the same image, showcasing diverse textures within the same terrain class. }
    \label{fig:knn}
\end{figure*}
\end{comment}

\begin{figure*}[htb]
\centering
 \begin{minipage}{.75\textwidth}
    \includegraphics[width=0.99\textwidth, height=2.5cm]{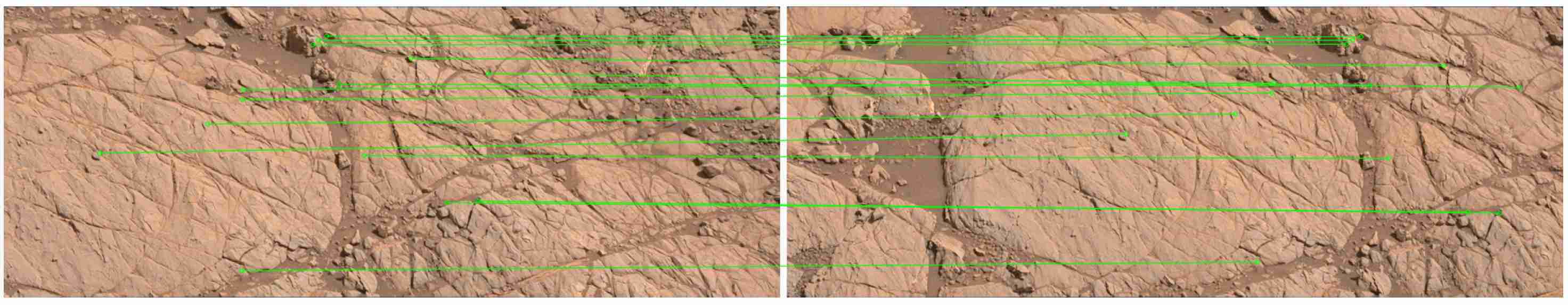}
    \subcaption{}
    \includegraphics[width=0.49\textwidth, height=2.5cm]{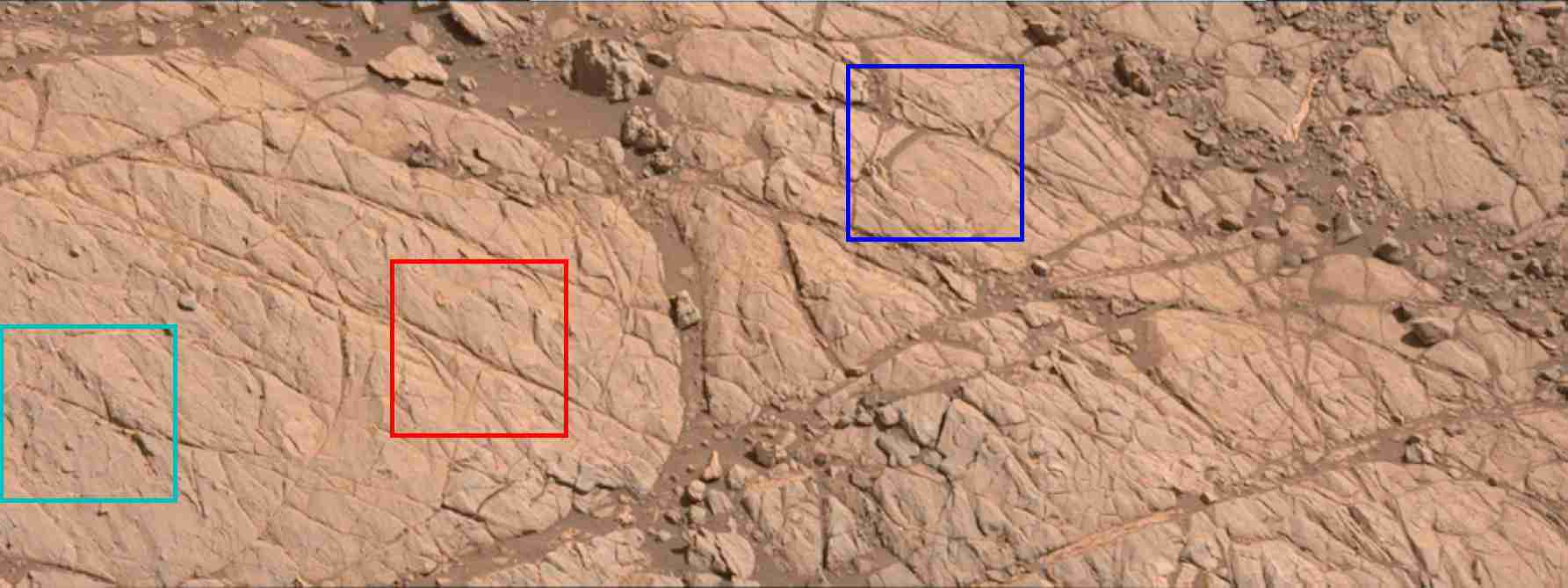}
    \includegraphics[width=0.49\textwidth, height=2.5cm]{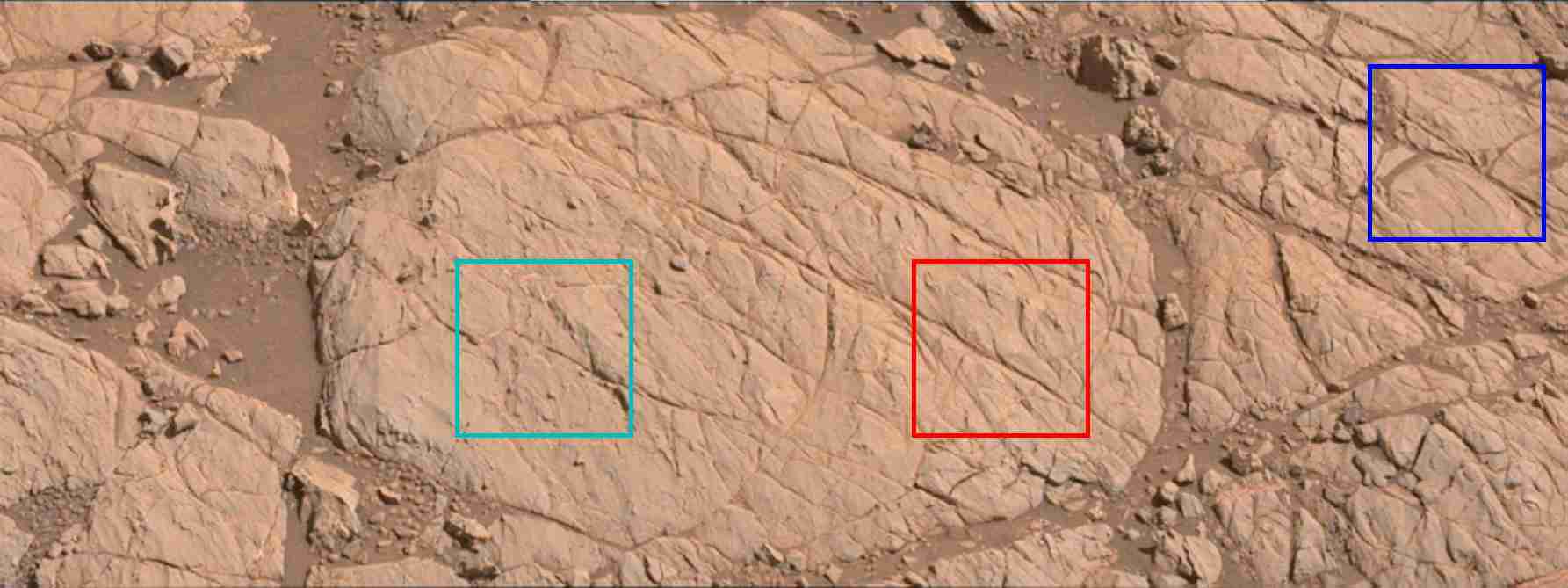}
    \subcaption{}
\end{minipage}
\begin{minipage}{.24\textwidth}
    \includegraphics[width=0.49\textwidth, height=1.9cm]{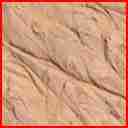}
    \includegraphics[width=0.49\textwidth, height=1.9cm]{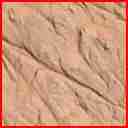}
    
    \includegraphics[width=0.49\textwidth, height=1.9cm]{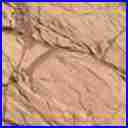}
    \includegraphics[width=0.49\textwidth, height=1.9cm]{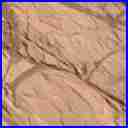}

    \includegraphics[width=0.49\textwidth, height=1.9cm]{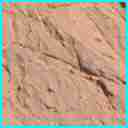}
    \includegraphics[width=0.49\textwidth, height=1.9cm]{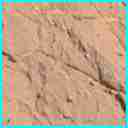}
    \subcaption{}
\end{minipage}
\centering
 \begin{minipage}{.75\textwidth}
    \includegraphics[width=0.99\textwidth, height=2.5cm]{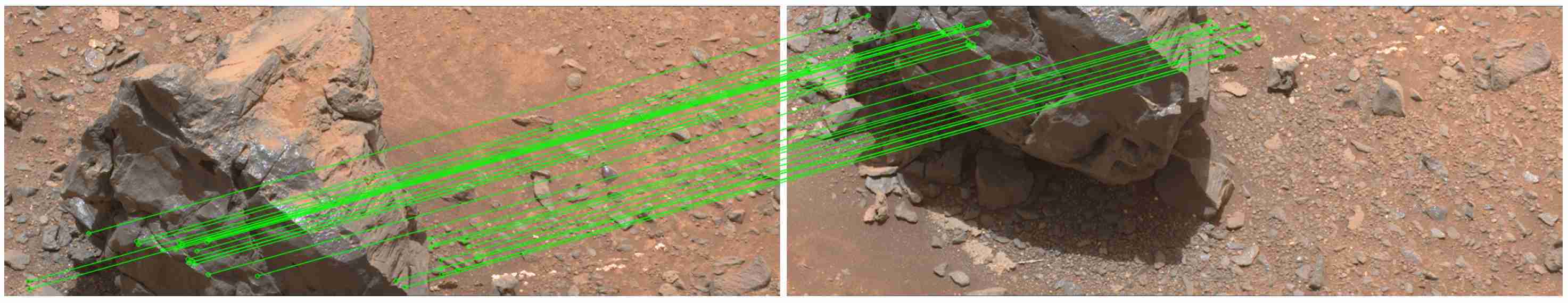}
    \subcaption{}
    \includegraphics[width=0.49\textwidth, height=2.5cm]{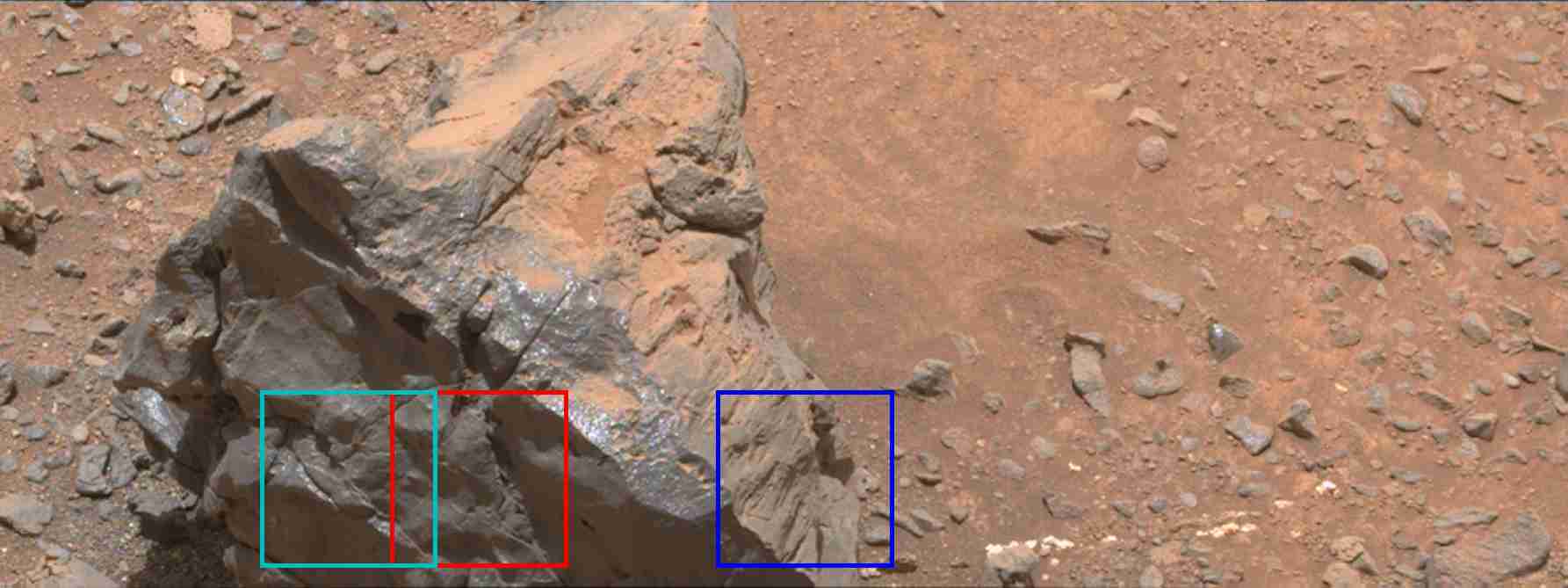}
    \includegraphics[width=0.49\textwidth, height=2.5cm]{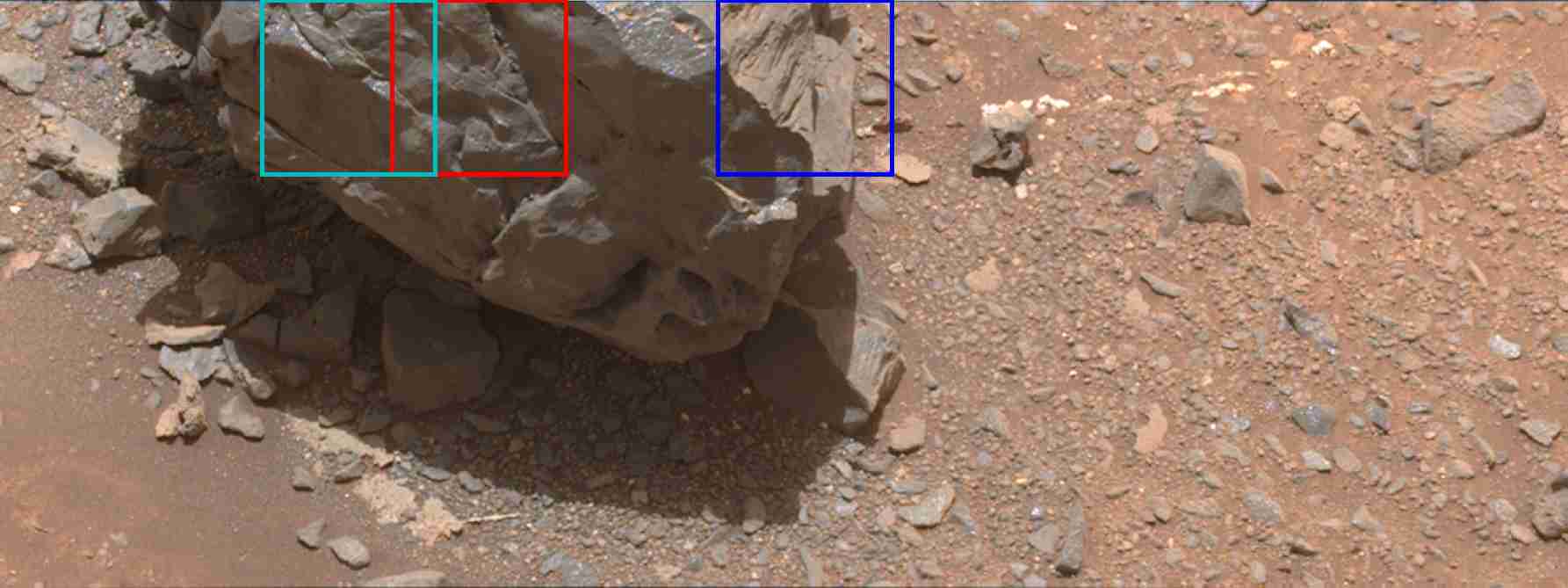}
    \subcaption{}
\end{minipage}
\begin{minipage}{.24\textwidth}
    \includegraphics[width=0.49\textwidth, height=1.9cm]{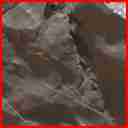}
    \includegraphics[width=0.49\textwidth, height=1.9cm]{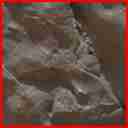}
    
    \includegraphics[width=0.49\textwidth, height=1.9cm]{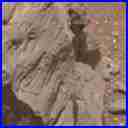}
    \includegraphics[width=0.49\textwidth, height=1.9cm]{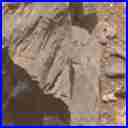}

    \includegraphics[width=0.49\textwidth, height=1.9cm]{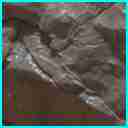}
    \includegraphics[width=0.49\textwidth, height=1.9cm]{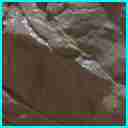}
    \subcaption{}
\end{minipage}
 
\caption{In subfigures (a) and (d), RSM image pairs illustrate terrain features identified by SIFT-detected keypoints highlighted with green lines. Subfigures (b) and (e) display three of the identified RSM image pairs with bounding boxes in red, blue, and cyan. Each row in subfigures (c) and (e) presents the enlarged version of pairs, ensuring consistent terrain characteristics across paired patches, as detailed in Section \cref{cc:RSM}.}
\label{fig:RSM}
\end{figure*}

\begin{figure*}[htb]

\includegraphics[width=\textwidth, height=6.5cm]{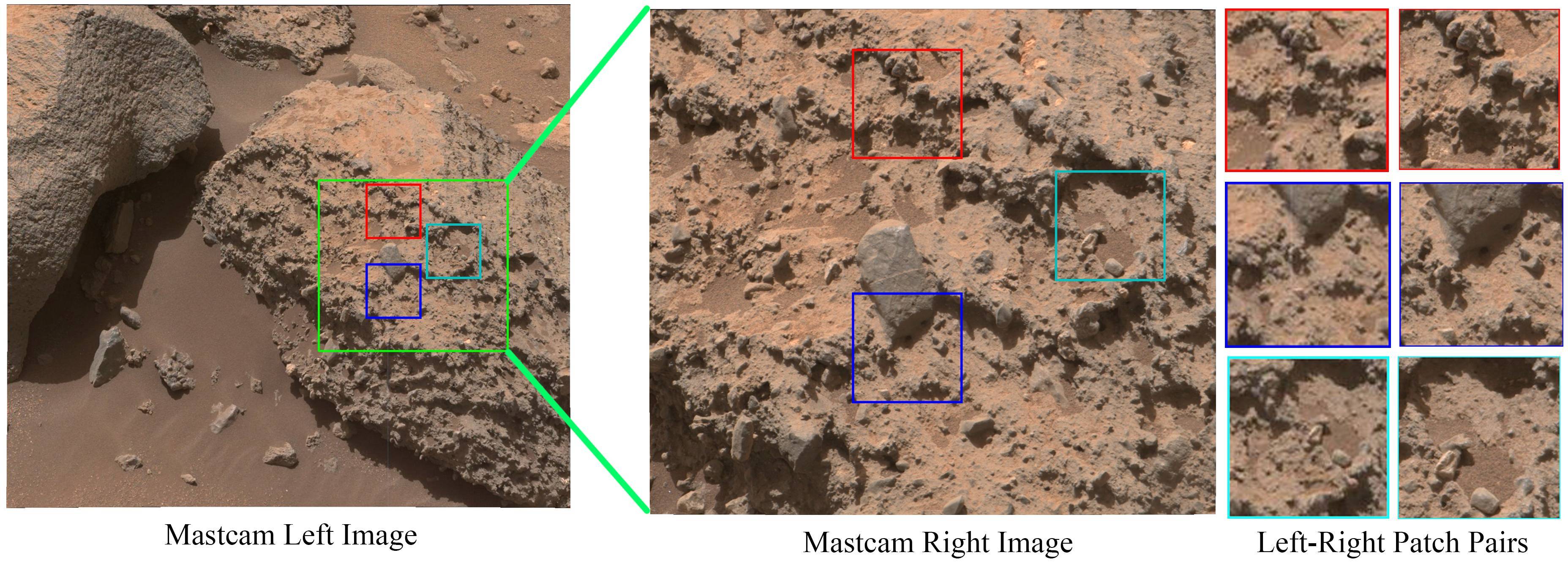}
\caption{Subfigures shows left-right image pairs with the right image localized on the left using green box, based on template matching algorithm. Each row in left-right patch pairs displays the matched patches from these regions, aligned using template matching techniques, as explained in Section \ref{cc:LR}.}
\label{fig:LR}
\end{figure*}

\begin{comment}
    
\begin{figure*}[htb]
\centering
 \begin{minipage}{.65\textwidth}
    \includegraphics[width=0.49\textwidth, height=5cm]{LRpairs/1408ML0068940010601914E01_DRCL_L.jpg}
    \includegraphics[width=0.49\textwidth, height=5cm]{LRpairs/1408ML0068940010601914E01_DRCL_R.jpg}
    \subcaption{}
\end{minipage}
\begin{minipage}{.33\textwidth}
    \includegraphics[width=\textwidth, height=2.5cm]{LRpairs/1408ML0068940010601914E01_DRCL_p1.jpg}
    \includegraphics[width=\textwidth, height=2.5cm]{LRpairs/1408ML0068940010601914E01_DRCL_p2.jpg}
    \subcaption{}
\end{minipage}
\centering
 \begin{minipage}{.65\textwidth}
    \includegraphics[width=0.49\textwidth, height=5cm]{LRpairs/1496ML0075680010603883E01_DRCL_L.jpg}
    \includegraphics[width=0.49\textwidth, height=5cm]{LRpairs/1496ML0075680010603883E01_DRCL_R.jpg}
    \subcaption{}
\end{minipage}
\begin{minipage}{.33\textwidth}
    \includegraphics[width=\textwidth, height=2.5cm]{LRpairs/1496ML0075680010603883E01_DRCL_p1.jpg}
    \includegraphics[width=\textwidth, height=2.5cm]{LRpairs/1496ML0075680010603883E01_DRCL_p2.jpg}
    \subcaption{}
\end{minipage}
 
\caption{Stereo Image Analysis and Patch Matching. Subfigures (a) and (c) show left-right image pairs with the right image localized on the left using cyan boxes, based on SIFT keypoint matching. Each row in subfigures (b) and (d) displays the matched patches from these regions, aligned using template matching techniques, as explained in Section \ref{cc:LR}.}
\end{figure*}

\end{comment}

To evaluate the effectiveness of the proposed deep constrained clustering algorithm, we conduct extensive experiments on the Curiosity dataset. We compare the clustering performance with previous deep clustering methods and evaluate both qualitatively, through visual inspection of the formed clusters, and quantitatively, using cluster evaluation metrics such as the DB-Index. The results demonstrate significant improvements in clustering performance, showcasing the potential of the proposed approach for robust Martian terrain recognition.

To summarize,  we propose Deep Constrained Clustering with Metric Learning (DCCML), an unsupervised clustering framework designed to improve Martian terrain recognition by integrating domain-specific constraints. The key contributions of our work are:

\begin{itemize}
    \item We introduce a constraint-guided clustering framework that enhances the robustness and consistency of terrain classification by incorporating spatial, depth, stereo, and RSM-based constraints.
    \item We develop a methodology for automatically generating pairwise constraints, ensuring that clustering decisions are informed by meaningful geological relationships rather than superficial visual similarities.
    %\item By integrating these constraints we achieve a 16.7\% increase in homogeneous clusters over previous methods on the Curiosity dataset, validated through DB-Index evaluation and qualitative analysis.
    \item We conduct extensive ablation studies to evaluate the effect of varying cluster counts, the contribution of individual constraints, and the performance of different network architectures (DEP and ViT).
\end{itemize}

The remainder of this paper is structured as follows: In \cref{ss:relatework}, we review related work in terrain recognition and self-supervised learning. \cref{ss:method} introduces the network architecture for deep constrained clustering, describes the training objective, and details the constraints incorporated in the clustering process. \cref{dataset} discusses the dataset used for experimentation and outlines the necessary preprocessing steps. We present experimental results and analysis in \cref{ss:res}, followed by the conclusion in \cref{ss:conclusion}, where we summarize our key contributions and discuss future research directions.

\section{Related work}
\label{ss:relatework}
\subsection{Self-Supervised Learning}

Self-supervised learning (SSL) is a widely studied paradigm within unsupervised representation learning, where models learn from automatically generated signals derived from the data itself, rather than requiring human-annotated labels \cite{simclr, moco, swav, byol, simsiam, barlow, colorization, jigsaw, rotation}. The core idea is to produce representations that embed semantically similar examples close to each other in the representation space. To this end, various SSL approaches have been developed, broadly categorized by their pretext tasks and optimization strategies.

Early discriminative SSL methods focused on predicting transformations applied to the data. For instance, rotation prediction \cite{rotation} trains models to identify the rotation angle of an image, solving jigsaw puzzles \cite{jigsaw} involves predicting the correct spatial arrangement of shuffled image patches, and patch location prediction \cite{doersch2016} tasks models with identifying the relative positions of image patches. Although these methods demonstrated the potential of self-supervised learning, their limitations in capturing high-level semantic features spurred the development of more advanced frameworks.

The advent of contrastive learning reshaped the landscape of SSL, introducing frameworks such as SimCLR \cite{simclr}, MoCo \cite{moco}, BYOL \cite{byol}, and SimSiam \cite{simsiam}. These methods generate multiple views of the same instance via data augmentation, aligning their representations while ensuring separation from other instances. SimCLR achieves this using strong augmentations (e.g., cropping, color distortion) and a contrastive loss to maximize agreement between positive pairs (augmented views of the same image) and minimize agreement with negatives (other instances). MoCo extends this by incorporating a memory bank of negative examples, while BYOL and SimSiam eliminate the need for negative pairs through momentum encoders or stop-gradient mechanisms, respectively, to prevent collapse. Barlow Twins \cite{barlow} further reduces the reliance on negative samples by leveraging redundancy reduction, aligning representations while minimizing cross-correlation between features. Similarly, DINO \cite{dino} employs a teacher-student framework, using self-distillation to align embeddings without explicit contrastive losses. While effective, these methods rely heavily on effective augmentation strategies. Augmentations suited for natural image datasets (e.g., random cropping, color jittering) can distort essential geological features in planetary data, leading to representations that fail to capture subtle but critical variations inherent to Martian terrains. Moreover, contrastive methods often focus on instance-level discrimination rather than encoding high-level semantic groupings, limiting their ability to capture intrinsic data structures.

In contrast, clustering-based methods inherently group similar data points, making them well-suited for tasks requiring natural semantic groupings without reliance on augmentations. Clustering-based SSL methods, such as DeepCluster \cite{deepcluster} and SwAV \cite{swav}, organize data into clusters and use the resulting pseudo-labels to supervise representation learning. However, their success depends on the quality of the clustering algorithm, which can be sensitive to variations in scale, illumination, and texture. To address these challenges, our approach enhances clustering-based SSL by introducing additional constraints that guide the clustering process. By learning a representation space where similar patches are grouped together, the model effectively captures meaningful geological classes without relying on superficial variations.

\subsection{Deep Constrained Clustering}
Traditional constrained clustering algorithms \cite{wagstaffconstrained, ccml, ccsa, probframework, XingNJR02} aim to improve clustering accuracy by incorporating pairwise constraints, such as must-link and cannot-link pairs, derived from ground truth labels in a semi-supervised setting. These methods typically operate directly on raw data or handcrafted features, which limits their ability to capture the complex structures inherent in high-dimensional, noisy, or non-linear datasets. Consequently, their performance often deteriorates when applied to such challenging scenarios.

Deep constrained clustering algorithms address these limitations by jointly learning feature representations and clustering assignments, enabling them to better capture the underlying structure of complex datasets \cite{zhang2019, hsu2016neural, shaham2018, guo2017improved, ren2022deep}. Most of these methods employ autoencoder-based architectures with specialized clustering losses. However, autoencoders are inherently optimized for reconstruction tasks, which may not ensure that the learned representations are suitable for effective clustering \cite{dec2016}. This misalignment between reconstruction objectives and clustering goals often results in suboptimal clustering performance.

In this work, we build upon a deep clustering framework \cite{deepcluster} and extend it with key enhancements developed in our prior studies \cite{tejas1, tejas2, tejasearth}. This framework follows an iterative clustering-refinement approach, where embeddings are alternately clustered and refined through metric learning to improve feature representation.

To further enhance clustering performance, we replace the standard Kmeans clustering module with Pairwise Confidence-Constrained Clustering (PCCClustering) \cite{PCCC}, which integrates pairwise constraints directly into the clustering process. Unlike traditional clustering methods that rely solely on geometric similarity in feature space, PCCClustering incorporates must-link and cannot-link constraints, ensuring that geologically meaningful relationships are preserved. To address the absence of labeled data for constraint formation, we automatically generate soft and hard constraints based on a combination of spatial proximity, depth similarity, and other contextual relationships, as detailed in \cref{subsec:genc}. By leveraging these constraints, the model becomes increasingly robust to natural variations in scale, illumination, and other non-informative or misleading factors. Consequently, the proposed approach produces semantically meaningful clusters that align with scientific classifications, enabling more accurate and geologically relevant terrain analysis.

\subsection{Martian Terrain recognition}

The classification of Martian terrains has predominantly relied on supervised learning methods that use annotated datasets for segmentation and navigation tasks. Prior works such as AI4Mars \cite{ai4mars} employ crowd-sourced annotations to categorize terrain into broad classes—soil, bedrock, sand, and big rock, facilitating terrain analysis and rover operations. Similarly, other supervised approaches \cite{labelmars, demud} focus on identifying specific geological features (e.g., layered rock, veins, or sand) aligned with distinct mission objectives. Despite these successes, such methods require extensive manual labeling, which is laborious and not sufficiently granular for detailed geological exploration.

To address these issues, researchers have explored self-supervised and unsupervised methods for terrain recognition. Goh et al. \cite{jplcontrastive}, for instance, introduced a semi-supervised framework for Mars terrain segmentation by pretraining a deep segmentation network on unlabeled data and fine-tuning it with a smaller set of labeled samples, demonstrating strong performance under limited annotation scenarios. Beyond semi-supervision, fully self-supervised and unsupervised strategies aim to reduce or eliminate dependence on large labeled datasets.

In our previous work \cite{tejas1}, we utilized an alternating clustering and classification strategy similar to DeepCluster \cite{deepcluster} to identify homogeneous clusters within Mars terrain images. Building on this, in DeepClustering using Metric Learning (DCML) \cite{tejas2} we leveraged a triplet network trained through iterative clustering to enhance cluster separability, achieving greater inter-class distinction and reduced intra-class variation. Despite these advancements, these methods faced limitations due to the shortcomings of conventional CNN architectures, which struggle to effectively capture the intricate and fine-grained characteristics of Martian terrain textures. To overcome these challenges, we incorporated the texture encoding module from \cite{tejasearth}, which introduced a texture-based clustering methodology paired with an exhaustive classification taxonomy for Curiosity Mastcam images. Later, in \cite{mixedpatch}, we introduced coarse segmentation and depth estimation techniques to filter out mixed terrain patches (where multiple terrains overlap) and distant patches (captured from far-away ranges). By discarding these noisy patches, the resulting clusters better represented single-terrain types, improving classification performance and enhancing scientific interpretability. Building on these advances, this paper introduces Deep Constrained Clustering with Metric Learning (DCCML), which incorporates pairwise “must-link” constraints derived from stereo-camera pairs, consecutive (RSM) frames, and spatially close patches of similar depth.  
These constraints improve cluster robustness by ensuring that patches exhibiting similar terrain characteristics are grouped together, despite natural variations in appearance, resulting in clusters that more faithfully represent the underlying geological structures. 

%%%%%%%%% METHODOLOGY %%%%%%%%% 
\section{Deep self-supervised texture recognition with constrained clustering}
\label{ss:method}
In this section, we first introduce the network architectures employed in our approach, as detailed in \cref{subsec:dep}. We then describe our self-supervised training objective in \cref{subsec:dcml}. Next, in \cref{subsec:cc}, we present our clustering algorithm, which incorporates soft and hard constraints. Finally, \cref{subsec:genc} outlines our method for generating constraints between image patches.

\subsection{Network Architectures}
\label{subsec:dep}

\begin{figure*}[htb]
    \centering
    \includegraphics[width=0.8\textwidth, height=9.5cm]{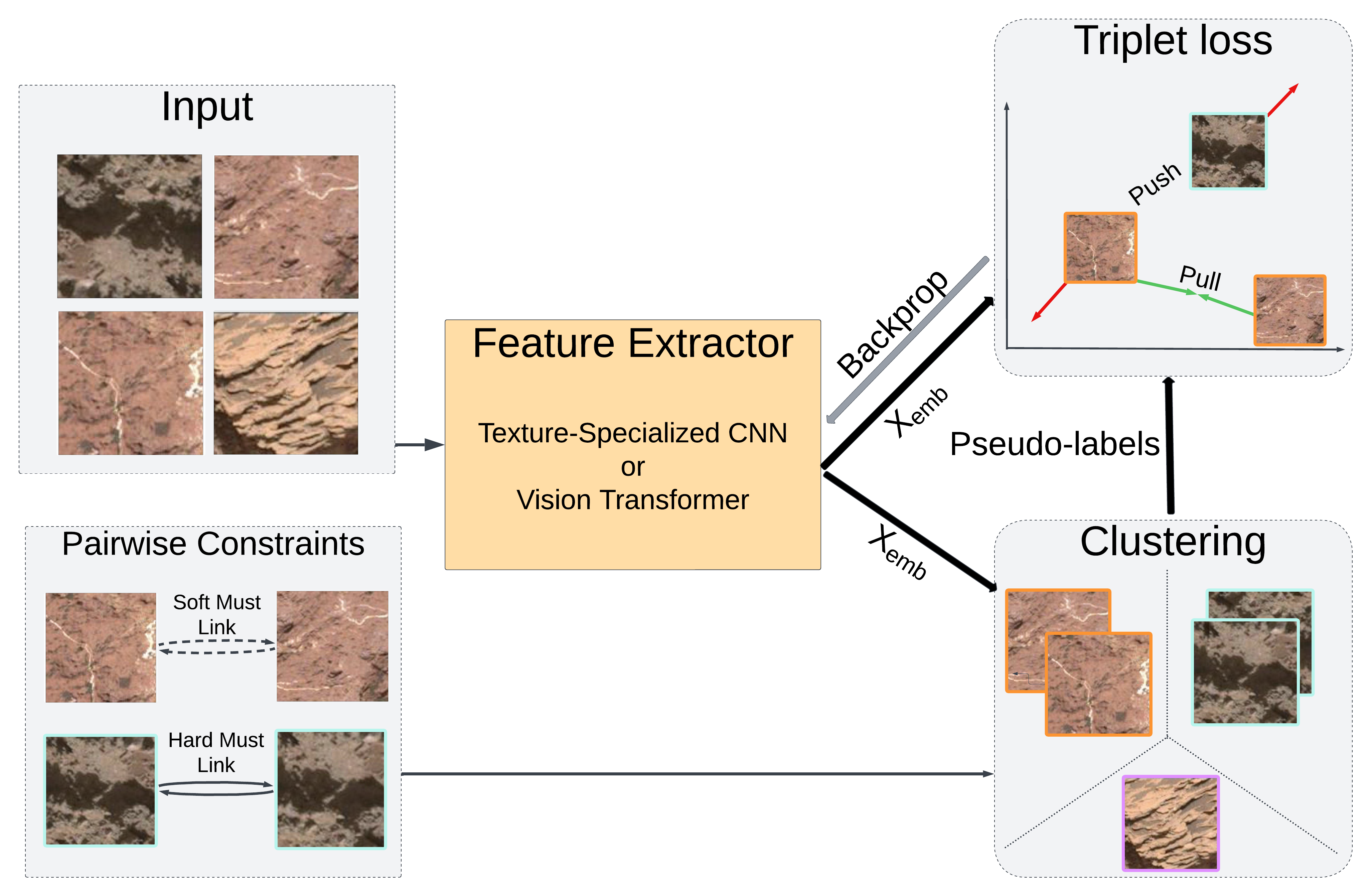}
    \caption{Overview of our approach: Deep Constrained Clustering with Metric Learning (DCCML) employs an iterative framework to refine feature embeddings and clustering assignments. The method integrates spatial and depth-based soft constraints from neighboring patches and hard constraints derived from stereo-camera pairs and consecutive RSM frames. This ensures the formation of coherent and semantically meaningful clusters for Martian terrain recognition.}

    \label{fig:feature_extraction}
\end{figure*}

Our methodology utilizes two distinct architectures to extract feature embeddings ($\mathbf{X_{emb}}$) from image patches ($\mathbf{I_p}$): a texture-specialized convolutional neural network called the Deep Encoding Pooling Network (DEP), which is based on the ResNet-18 architecture \cite{gtos-dep}, and a Vision Transformer (ViT) \cite{vit}. As illustrated in Figure~\ref{fig:feature_extraction}, these architectures serve as the backbone for generating embeddings used in the subsequent constrained clustering process.

In the DEP architecture, ResNet-18 processes input patches to generate intermediate feature representations. These features are refined through two additional layers: a texture encoding layer \cite{depten}, which captures local texture patterns, and a global average pooling (GAP) layer, which aggregates information across the entire patch. The outputs of these layers are combined using bilinear pooling \cite{bilinear}, producing the final embedding, $\mathbf{X_{emb}}$, which represents both localized details and broader spatial information.

The ViT architecture, on the other hand, processes image patches through a series of self-attention mechanisms and feed-forward layers to directly compute $\mathbf{X_{emb}}$. By leveraging self-attention, the ViT effectively models relationships between image regions, capturing both fine-grained local details and broader contextual patterns within the input.

The embeddings generated by either architecture are subsequently fed into the constrained clustering framework. By utilizing these architectures, we assess the robustness and adaptability of our clustering methodology. 

\subsection{Deep constrained clustering using metric learning}
\label{subsec:dcml}
Our approach combines deep constrained clustering with metric learning in an iterative framework to refine the feature embedding space and incorporate pairwise constraints. The process begins with clustering the embeddings ($\mathbf{X_{emb}}$) extracted from the network, where the resulting cluster assignments are treated as pseudo-labels. These pseudo-labels are then used to generate triplets for metric learning, each consisting of an anchor, a positive, and a negative example. Positive examples are selected from the same cluster as the anchor, while negative examples are chosen from different clusters.

To optimize the embedding space, we use triplet loss \cite{tejas2} as the objective function. This loss minimizes the distance between anchor-positive pairs while maximizing the distance between anchor-negative pairs in the embedded space. Our triplet sampling strategy follows the method described in \cite{tejas2, tejasearth} to ensure efficient and representative triplet selection.

The clustering process incorporates pairwise constraints, enforcing must-link and cannot-link relationships between data points. The mechanism for integrating these constraints is detailed in \ref{subsec:cc}. 

\subsection{Constraint-Based Clustering}
\label{subsec:cc}
We adopt the Pairwise-Confidence-Constrained Clustering (PCCClustering) approach \cite{PCCC} to enhance the clustering process by incorporating pairwise constraints derived from spatial, depth, and stereo-based relationships in the dataset. These constraints ensure that clustering assignments respect both known relationships in the data and intrinsic patterns within the feature space.

For the hard must-link constraints, we utilize RSM pairs and Left-Right (LR) pairs from the dataset. RSM pairs connect patches from consecutive images, ensuring that patches representing the same terrain are grouped together despite minor positional shifts. Similarly, LR pairs link patches from stereo images captured simultaneously by the Curiosity rover’s left and right cameras. The methodology for generating these RSM and LR pairs is described in the following subsection \ref{cc:RSM} and \ref{cc:LR}.

In addition to hard constraints, soft must-link constraints are introduced to capture probabilistic relationships between neighboring patches within the same image. These relationships are quantified using similarity scores derived from spatial proximity and depth differences. Each similarity score serves as a confidence value, which directly influences the clustering process.

The confidence level associated with each constraint determines the extent to which it is enforced. Hard must-link constraints are enforced absolutely, meaning any violation significantly impacts clustering accuracy. Conversely, soft must-link constraints are characterized by flexible confidence levels derived from similarity scores. Constraints with higher confidence values are more strictly enforced, emphasizing stronger relationships, while those with lower confidence values are relaxed, allowing the algorithm to optimize the overall clustering objective.

By integrating both hard and soft constraints, the PCCClustering algorithm iteratively refines clustering assignments, preserving strong relationships dictated by hard constraints while balancing softer, similarity-driven relationships. This approach results in more accurate and semantically meaningful clusters that align closely with the underlying data structure.

\subsection{Generating Constraints between Neighbouring Patches}
\label{subsec:genc}
In this section, we explain the procedure to obtain the pairwise constraints.  
\subsubsection{Spatial Constraints}
Spatial constraints are derived based on the proximity of patches within the same image. Let the center coordinates of patch $P_1$ be $(i_1, j_1)$ and those of patch $P_2$ be $(i_2, j_2)$, where the image dimensions are $m \times n$. The spatial distance, $d_{\text{spatial}}$, between these two patches is calculated as the squared Euclidean distance:

\begin{equation}
d_{\text{spatial}} = (i_1 - i_2)^2 + (j_1 - j_2)^2
\end{equation}

This measure quantifies how close or far apart the patches are within the image. Proximity is an essential factor for establishing soft must-link constraints, as neighboring patches are more likely to represent similar terrain features. However, spatial proximity alone can be insufficient because Martian terrain images often contain features that are spatially adjacent but belong to different geological contexts due to their three-dimensional distribution.

\subsubsection{Depth Constraints}
Depth constraints extend the spatial relationships by incorporating information about the relative depths of the patches. Using a depth map generated by the MiDaS algorithm \cite{midas}, we compute the depth differences between corresponding pixels in patches $P_1$ and $P_2$. Let the depth values of the pixels in patch $P_1$ be ${d_{1,1}, d_{1,2}, \dots, d_{1,N}}$ and in patch $P_2$ be ${d_{2,1}, d_{2,2}, \dots, d_{2,N}}$, where $N$ is the number of pixels in each patch. The depth distance, $d_{\text{depth}}$, is defined as the mean squared difference:

\begin{equation}
d_{\text{depth}} = \frac{1}{N} \sum_{n=1}^{N} (d_{1,n} - d_{2,n})^2
\end{equation}

This metric evaluates the similarity in depth information, which is critical for distinguishing patches that are spatially close but represent different terrain types. For example, a rock patch and a nearby soil patch may appear adjacent in the image but have significantly different depths. Incorporating depth ensures that patches representing terrain features at different ranges are not mistakenly grouped solely based on spatial proximity, thereby improving the accuracy and coherence of the clusters.

\subsubsection{Soft Must-Link Constraints}
To integrate spatial and depth information into the clustering process, we define soft must-link constraints through a similarity measure, $sim$, which combines the spatial and depth distances:

\begin{equation}
sim = \alpha \cdot \exp\left(-\frac{d_{\text{spatial}}}{2 \cdot \sigma_{\text{spatial}}^2}\right) + \beta \cdot \exp\left(-\frac{d_{\text{depth}}}{2 \cdot \sigma_{\text{depth}}^2}\right)
\end{equation}

Here, $\sigma_{\text{spatial}}$ and $\sigma_{\text{depth}}$ are hyperparameters that control the relative influence of spatial and depth components. For our experiments, we set $\alpha = 0.5$ and $\beta = 0.5$ to ensure equal contribution from both spatial and depth similarity in defining soft constraints. Further, $\sigma_{\text{spatial}} = 512$ was chosen to allow each patch to connect to approximately $14$ spatially neighboring patches, ensuring local connectivity while avoiding excessive linking. $\sigma_{\text{depth}} = 6$ was determined based on the standard deviation of depth variations across multiple Martian terrain images, maintaining meaningful constraints that reinforce terrain continuity.

To establish soft must-link constraints, we apply a $0.7$ similarity threshold, discarding lower-scoring pairs. This threshold was selected through Davies-Bouldin Index (DBI) evaluation at $150$ clusters, where clustering performance was assessed across similarity thresholds ranging from $0.4$ to $0.9$ in increments of $0.1$. Lower thresholds over-connect patches, while higher thresholds limit the use of existing beneficial geological pairs, making constrained clustering behave more like K-means. After applying this threshold, each patch forms an average of $8$ valid soft must-link connections.

\cref{fig:NebrSim} illustrates this concept, where terrain images are shown alongside extracted patches, demonstrating diverse textures within the same geological class. The patches highlighted within each terrain image emphasize local connectivity, ensuring that clustering assignments capture meaningful geological structures while avoiding excessive linking across distinct terrain types.

Further, these similarity scores are used as confidence levels for the soft constraints in the PCCClustering algorithm, where higher scores mean the algorithm prioritizes satisfying those constraints, while lower scores provide flexibility to occasionally violate them for a better overall clustering objective. 

\subsection{Establishing Constraints between Left and Right Image Patches}
\label{cc:LR}
The Curiosity rover's Mastcam is equipped with two camera systems: a $34 mm$ focal length camera for wider field-of-view imaging and a $100 mm$ focal length camera for zoomed-in, high-resolution imaging. These cameras capture the same scene simultaneously, providing complementary perspectives of the Martian terrain. Smaller patches of $128X128$ pixels from the $34 mm$ camera and larger patches of $256X256$ pixels from the $100 mm$ camera are extracted using a sliding window approach with a $50\%$ stride. This patch extraction  strategy is discussed later in \cref{subsec:sodata}.

To establish correspondences between the left and right image pairs, we employ the Scale-Invariant Feature Transform (SIFT) algorithm, which is widely recognized for its robustness to scale and rotation changes \cite{Lowe}. SIFT detects keypoints and computes descriptors for both the left and right stereo images. The keypoints detected in the left and right images are denoted as $kp_{\text{left}}$ and $kp_{\text{right}}$ respectively, with descriptors represented as $desc_{\text{left}}$ and $desc_{\text{right}}$. 

Keypoint correspondences are determined by calculating the descriptor distances for each keypoint pair using the Brute-Force Matcher algorithm. This algorithm identifies pairs of keypoints with the smallest descriptor distances as matches that form a set of matched keypoints, $m_{\text{SIFT}}$. This matching process is a standard approach in stereo vision for establishing correspondences between image pairs \cite{Hartley2004}. 

Following keypoint matching, we extract the region of interest (ROI) around the matched keypoints from the left image, denoted as $ROI_{\text{left}}$. Subsequently, template matching is employed to pinpoint the exact location of the corresponding region in the right image. This involves searching within the $ROI_{\text{left}}$ for the best matching area of the right image. \cref{fig:LR} illustrates the left-right image matching process, where corresponding patches identified using template matching are visualized. Each row in the rightmost subfigure of \cref{fig:LR} shows the matched patch pairs aligned using this approach.

Once the right image is localized within the left image, the patches within these regions are mapped to establish hard must-link constraints between them. The algorithm considers these constraints as strict requirements that must be satisfied when assigning instances to clusters. These hard constraints are crucial for addressing the issue of clustering left and right pairs into separate groups due to inherent resolution differences and potential blurring effects. 

\subsection{Establishing Constraints between RSM Pairs}
\label{cc:RSM}

To generate additional must-link constraints, we utilize image pairs that share identical metadata attributes—location (site), rover driving path (drive), and camera orientation (pose)—but differ by two in their RSM (Rover State Machine) counts. The RSM count, recorded in the Planetary Data System (PDS) metadata serves as a sequential identifier for images captured during the rover's operations.

Image pairs with a two-RSM count difference typically capture views of the same terrain features. However, these features often appear with slight translations, variations in scale, illumination changes, and minor rotations between the two images. To address these variations, we adapt the SIFT-based matching methodology described in \cref{cc:LR} to this specific context.

Using SIFT \cite{Lowe}, keypoints are detected and descriptors are computed for both images in the RSM pair. The patches containing the highest number of matched keypoints are selected to ensure that corresponding terrain features are correctly paired. The matched patches are used to establish hard must-link constraints between the corresponding regions of the RSM image pairs. These constraints are crucial to ensure consistency in clustering assignments across images that capture the same terrain under slightly different conditions.

\cref{fig:RSM} illustrates this process, where subfigures (a) and (d) highlight SIFT-detected keypoints using green lines across RSM image pairs. The identified corresponding patches are marked with bounding boxes in red, blue, and cyan in subfigures (b) and (e). The rightmost subfigures (c) and (f) display enlarged views of these paired patches, ensuring terrain characteristics remain consistent despite minor viewpoint changes. These visualizations confirm that our matching process effectively aligns terrain regions, forming reliable must-link constraints.

%%%%%%%%% DATASET %%%%%%%%% 
\section{Dataset}
\label{dataset}

\subsection{Sourcing data}
\label{subsec:creation}
% Data curation (train/test split, etc.) and --> 60/40 image split pre-processing
\label{subsec:sodata}
Our Curiosity dataset consists of DRCL (decompressed, radiometrically calibrated, color corrected, and geometrically linearized) images \cite{malin2017mars} acquired by the MSL Mast cameras between sol 1 and sol 3200. Crucially, this dataset is unique in that it is the only one available with an expert-established taxonomy for fine-grained analysis of Martian terrains, providing a robust benchmark for evaluating our clustering results. Following the settings in \cite{tejas2}, we focus exclusively on terrain images, removing those containing rover parts, sky, and other irrelevant features. To minimize scale disparities in terrain features, we limit our analysis to images with a target distance of less than $15m$ from the rover. This filtering process results in a dataset of approximately $\sim35,000$ images of size typically $1200\times1600$ pixels. 

As Martian terrain images often contain multiple terrain classes, we extract smaller patches of size $128 \times 128$ and $256 \times 256$ using a sliding window with a 50\% stride. These patch sizes correspond to the focal lengths of the "left eye" and "right eye" cameras on the Curiosity rover. The extraction yields a total of $\sim3.6M$ patches.

\subsection{Preprocessing}
\label{subsec:preprocessing}

To refine the dataset and ensure meaningful clustering, we implement a two-stage patch preprocessing pipeline: (1) Mixed Patch and Distant Patch Removal, which eliminates patches containing multiple terrain types or those captured at large distances that lack sufficient resolution \cite{mixedpatch}, and (2) Soil and Pebble Patch Elimination, which filters patches of lower scientific interest to focus on geologically significant rock formations.

\subsubsection{Mixed Patch and Distant Patch Removal}
\label{subsec:mixed_distant_removal}

The first stage, Mixed Patch and Distant Patch Removal, begins with segmentation-based filtering to identify and eliminate mixed patches containing multiple terrain types. We employ Grounding Dino \cite{grounding-dino}, a pre-trained object detection model optimized for open-set detection, to localize rock structures within images. Given the textual prompt "rock", the model generates bounding boxes around detected rock regions, serving as region proposals for segmentation, as illustrated in \cref{fig3}a.

To segment individual rock formations, the bounding boxes are passed as prompts to the Segment Anything Model (SAM) \cite{SAM}, a state-of-the-art segmentation framework. As depicted in \cref{fig3}b, this step produces preliminary segmentation masks outlining rock regions. However, segmentation accuracy is affected by factors such as variations in rock texture, shape, and occlusions, occasionally leading to misclassifications.

To refine segmentation accuracy, we introduce a dedicated segmentation network that extends the SAM encoder with a custom decoder. The SAM encoder extracts feature representations, while the custom decoder, comprising convolutional transpose layers for upsampling, normalization layers for stability, and dropout layers for regularization generates refined segmentation masks. A self-training approach is applied, where initial SAM-generated masks serve as pseudolabels to fine-tune the decoder.

Following segmentation, patches are filtered based on terrain homogeneity. If a patch’s segmentation mask, as shown in \cref{fig3}c, contains fewer than $70\%$ of pixels from a single terrain class, it is classified as mixed and removed from the dataset, following \cite{mixedpatch}.

In addition to segmentation-based filtering, we also remove distant patches, which lack the resolution necessary for fine-grained geological analysis. Terrain patches with an absolute depth greater than 10m are discarded based on depth maps generated by \cite{midas}, following the methodology in \cite{mixedpatch}. After applying these preprocessing steps, the Curiosity dataset is reduced to approximately 2.6M patches.

\subsubsection{Soil and Pebble Patch Elimination}

\begin{figure*}[htb]
\centering
\includegraphics[width=\textwidth, height=14cm]{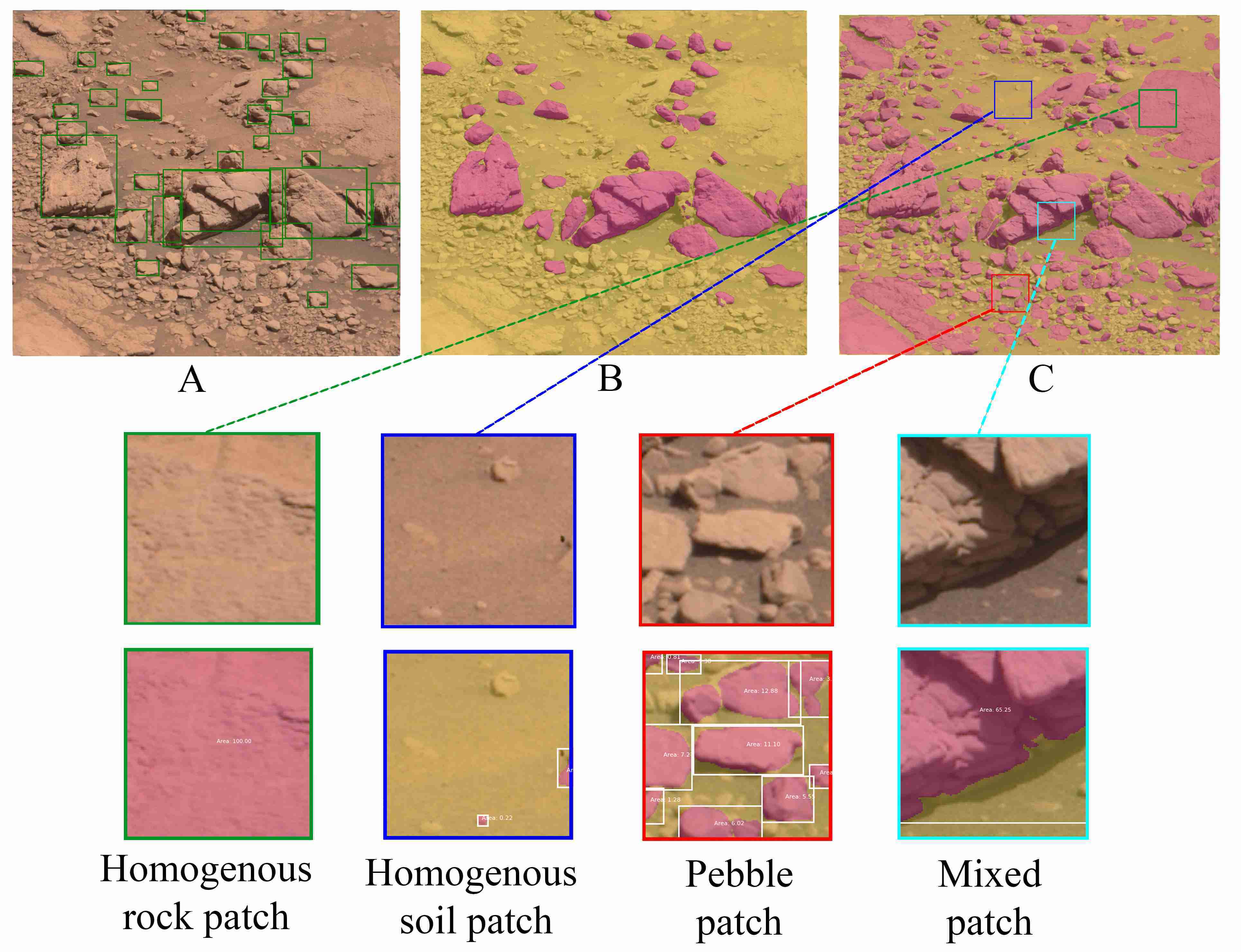}

\caption{Three subfigures (A, B, C) illustrating the process of coarse rock segmentation. (A) Bounding boxes generated using Grounding Dino for object detection. (B) Segmentation maps produced using SAM with the bounding boxes as prompts. (C) Refined segmentation masks following self-training, along with the localization of patches for further processing. Additionally, four extracted patches from (C) are shown, where the blue-bordered patch represents soil, the green-bordered patch corresponds to rock, the red-bordered patch denotes a pebble patch, and the cyan-bordered patch indicates a mixed terrain patch.}
\label{fig3}
\end{figure*}

Beyond filtering mixed and distant patches, we further refine the dataset by eliminating soil and pebble patches, as these terrains are abundant but of lower scientific interest. To achieve this, we analyze connected components within the refined segmentation maps, as shown in \cref{fig3}c. A patch is classified as soil if more than $70\%$ of its pixels are labeled as soil. Similarly, a patch is categorized as pebbly if it contains more than 10 connected components, indicative of fragmented terrain structures.

\Cref{fig3} presents four extracted patches representing different terrain types: a homogeneous soil patch (blue-bordered) where more than $70\%$ of pixels are labeled as soil, a homogeneous rock patch (green-bordered) dominated by a single rock structure, a pebble patch (red-bordered) where individual pebbles are detected as separate connected components, and a mixed patch (cyan-bordered) containing both rock and soil regions.

\subsection{Unbiased data for evaluation}
 Later, we split our data into train and test sets according to the procedure mentioned in \cite{tejasearth}. This approach mitigates potential distribution mismatches between the training and test sets. In particular, patches within the training set are chosen from the left $60\%$ portion of the given image, while patches from the remaining $40\%$ of the image are designated for the test set. For experiments involving terrain patch retrieval, patches acquired from the same \textit{site} and \textit{drive} of the rover are excluded from the evaluation results \cite{tejasearth}.
\begin{figure*}[t]
    \centering
    \includegraphics[width=0.45\textwidth]{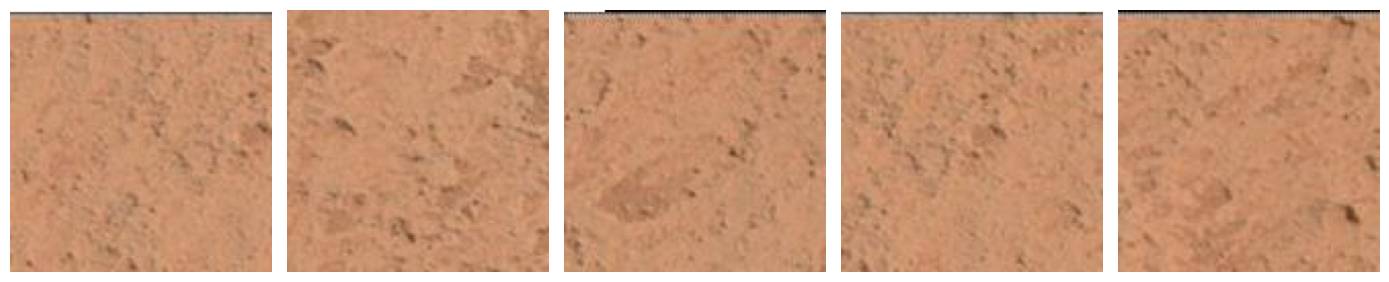}\hspace{0.05\textwidth}
    \includegraphics[width=0.45\textwidth]{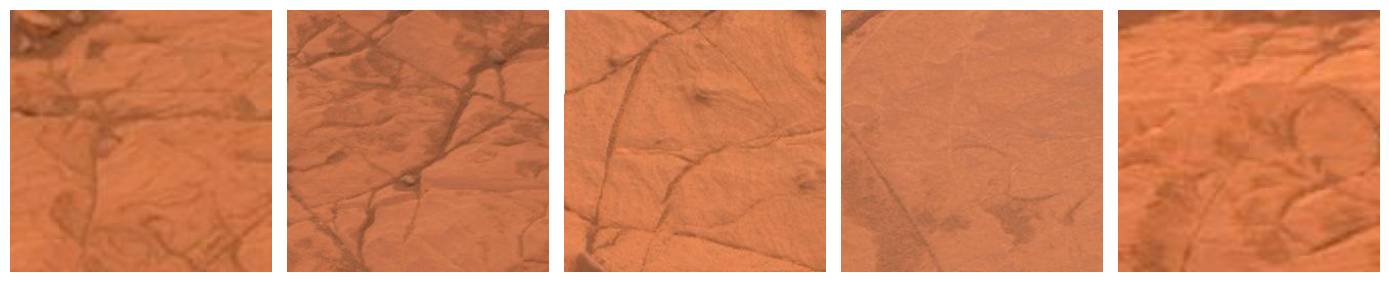}\hspace{0.05\textwidth}
    \includegraphics[width=0.45\textwidth]{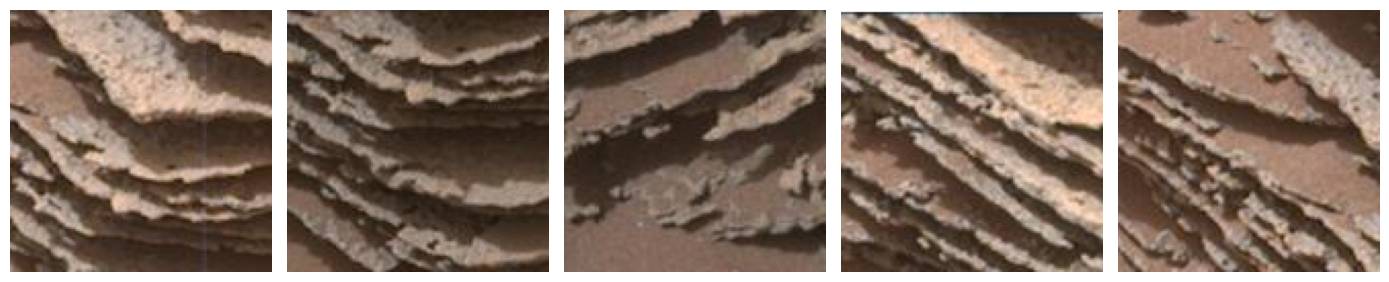}\hspace{0.05\textwidth}
    \includegraphics[width=0.45\textwidth]{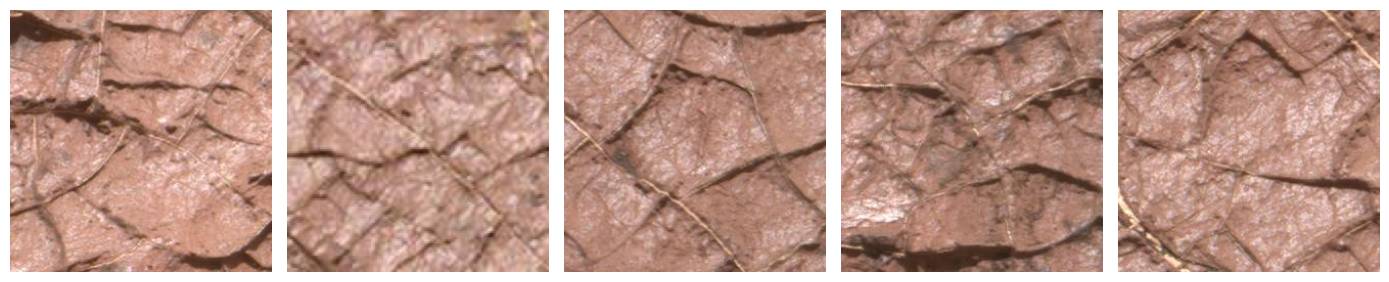}\hspace{0.05\textwidth}
    \includegraphics[width=0.45\textwidth]{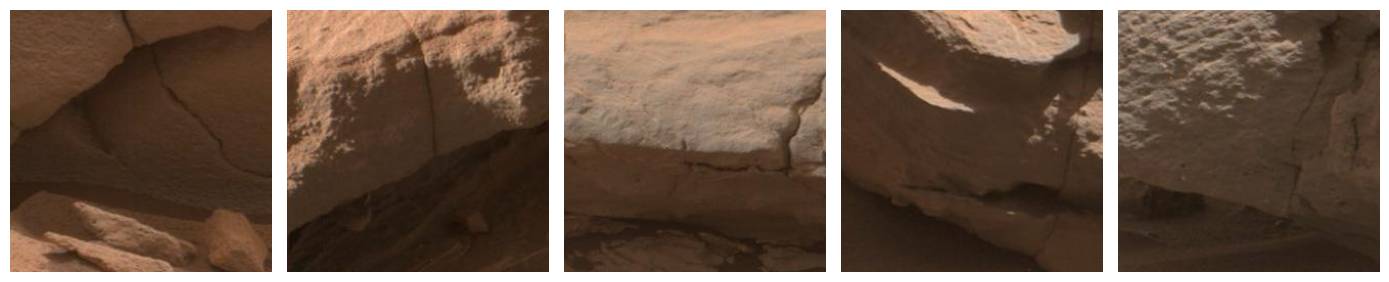}\hspace{0.05\textwidth}
    \includegraphics[width=0.45\textwidth]{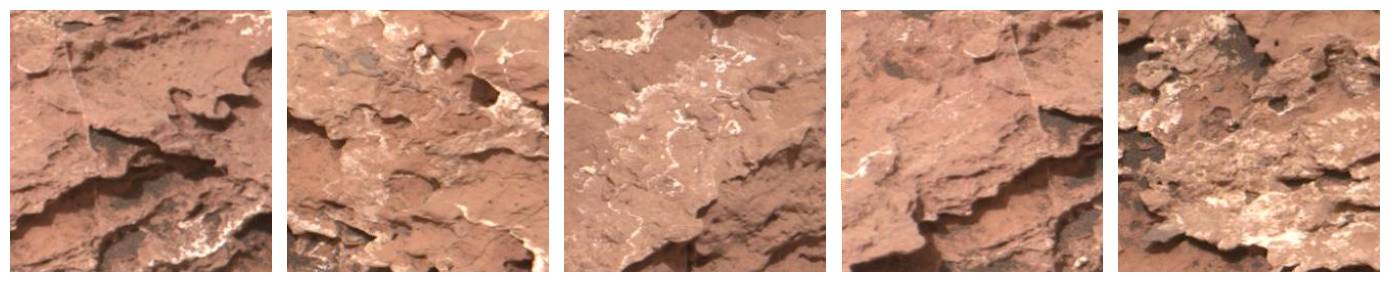}\hspace{0.05\textwidth}
    \includegraphics[width=0.45\textwidth]{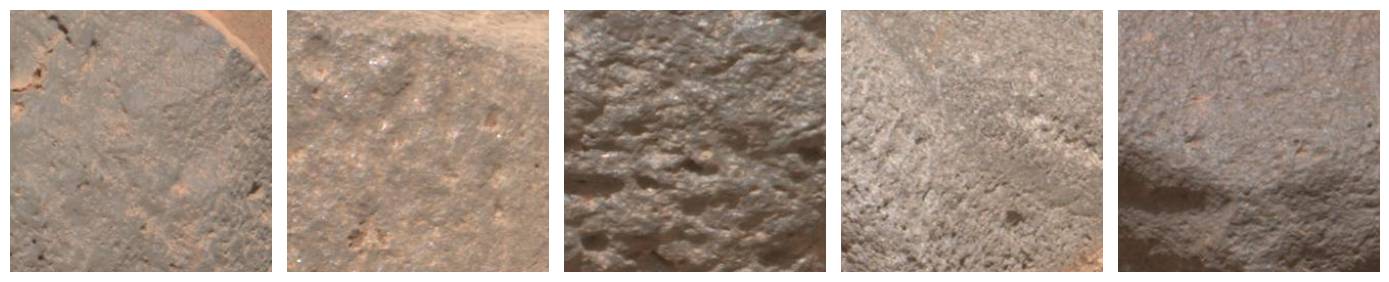}\hspace{0.05\textwidth}
    \includegraphics[width=0.45\textwidth]{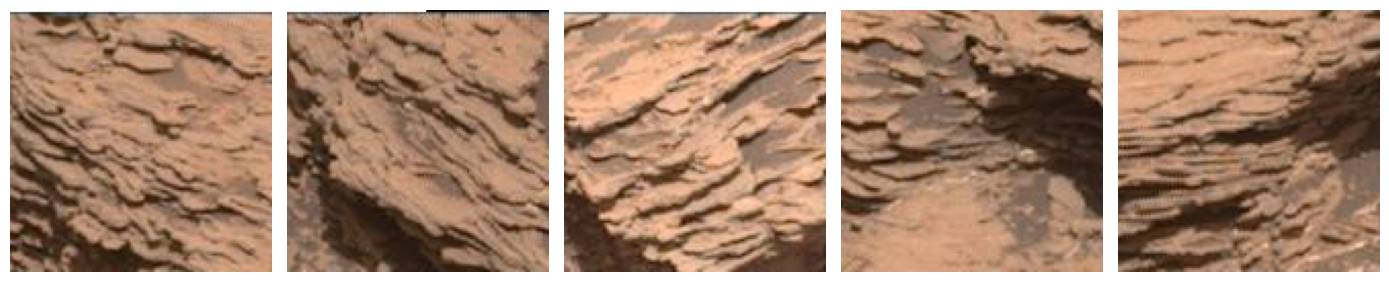}\hspace{0.05\textwidth}
    \includegraphics[width=0.45\textwidth]{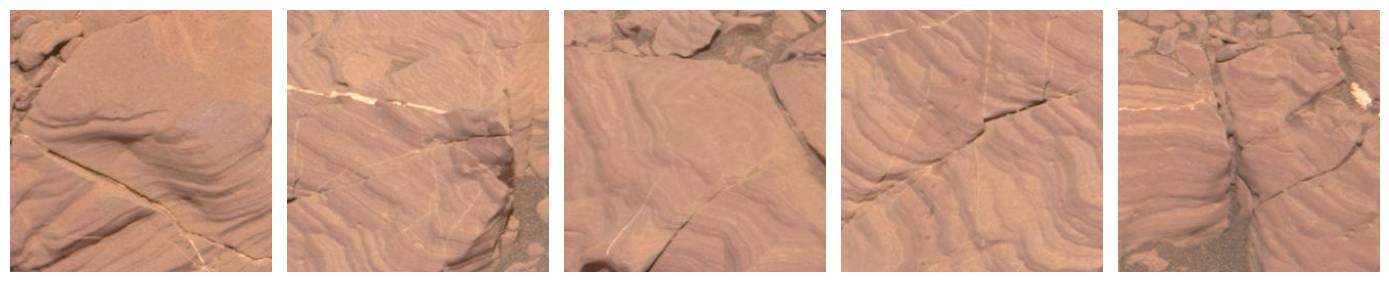}\hspace{0.05\textwidth}
    \includegraphics[width=0.45\textwidth]{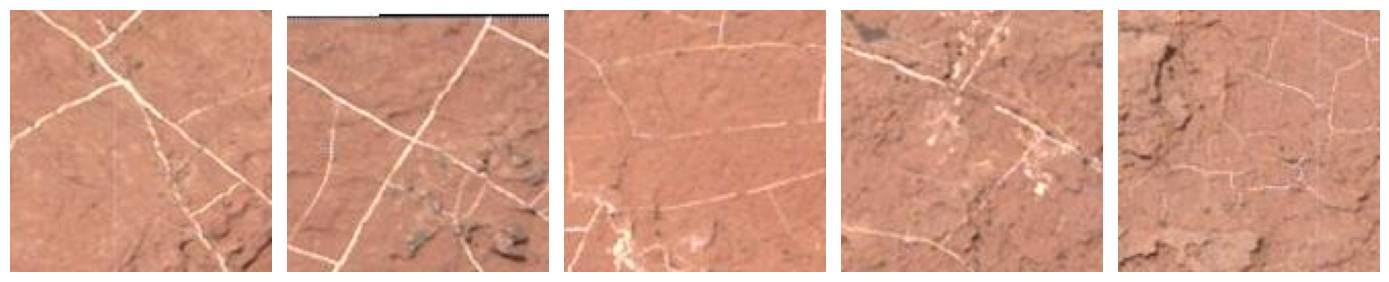}\hspace{0.05\textwidth}
    \includegraphics[width=0.45\textwidth]{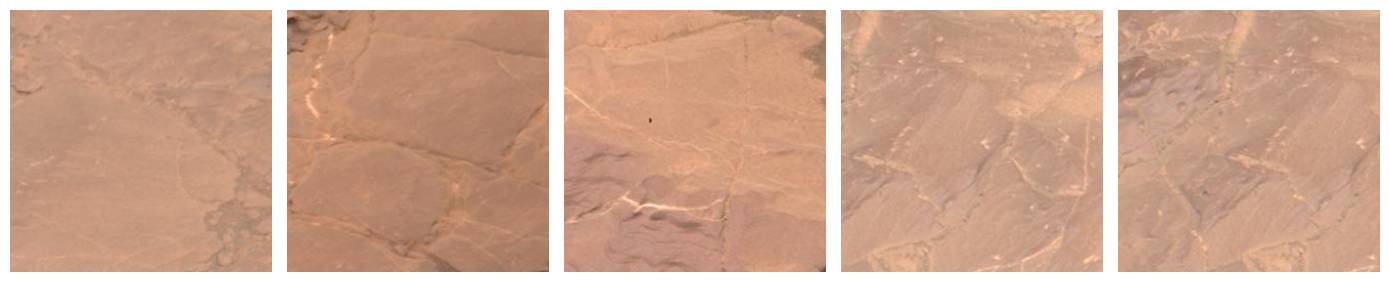}\hspace{0.05\textwidth}
    \includegraphics[width=0.45\textwidth]{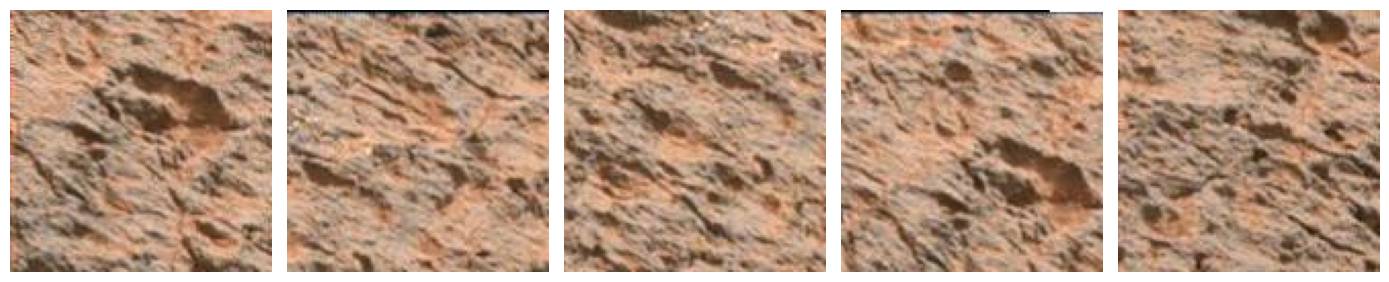}\hspace{0.05\textwidth}
    \includegraphics[width=0.45\textwidth]{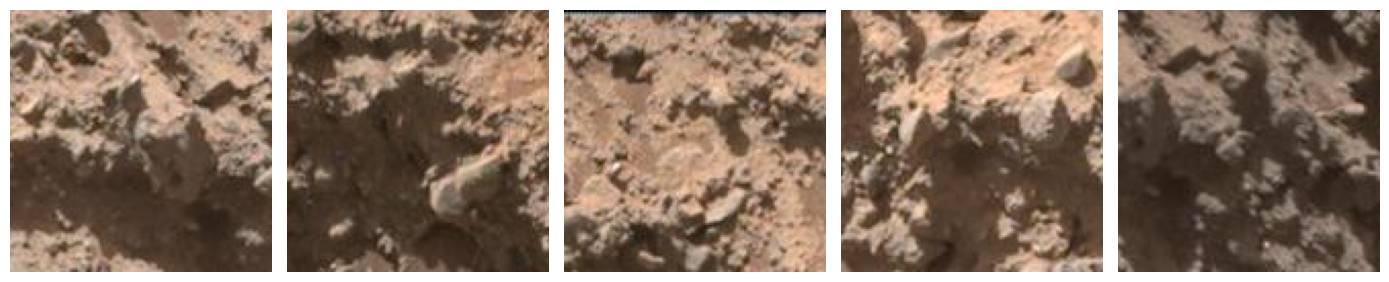}\hspace{0.05\textwidth}
    \includegraphics[width=0.45\textwidth]{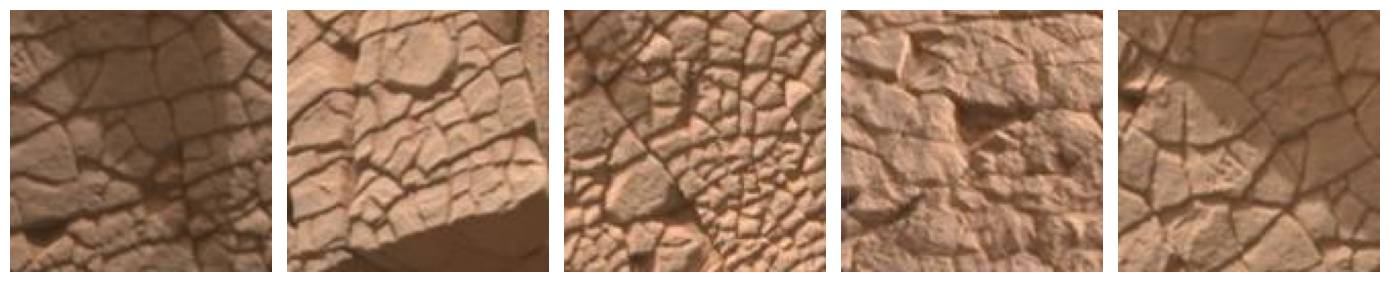}\hspace{0.05\textwidth}
    \includegraphics[width=0.45\textwidth]{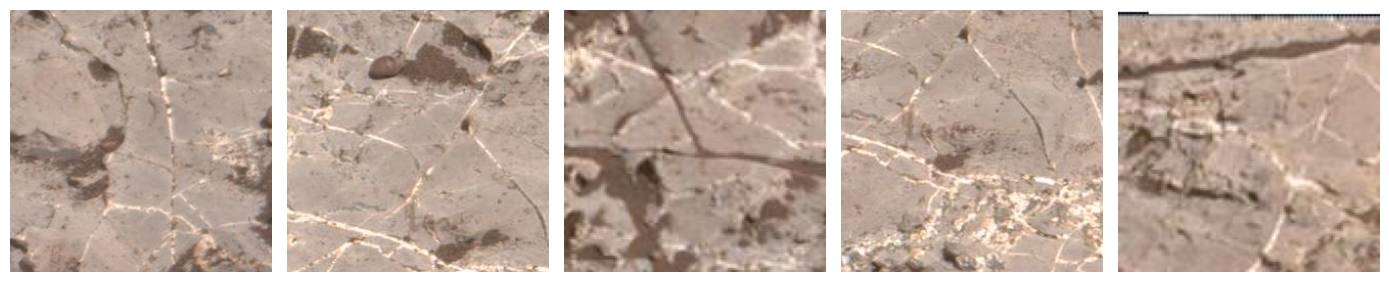}\hspace{0.05\textwidth}
    \includegraphics[width=0.45\textwidth]{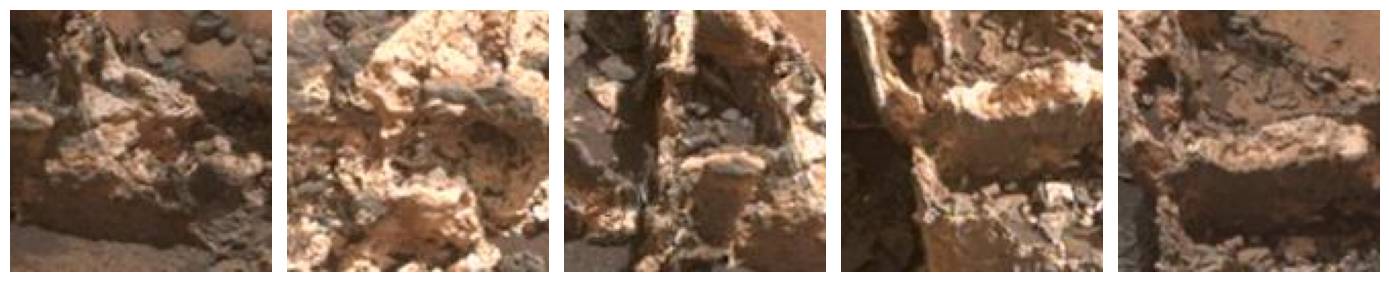}\hspace{0.05\textwidth}
    \includegraphics[width=0.45\textwidth]{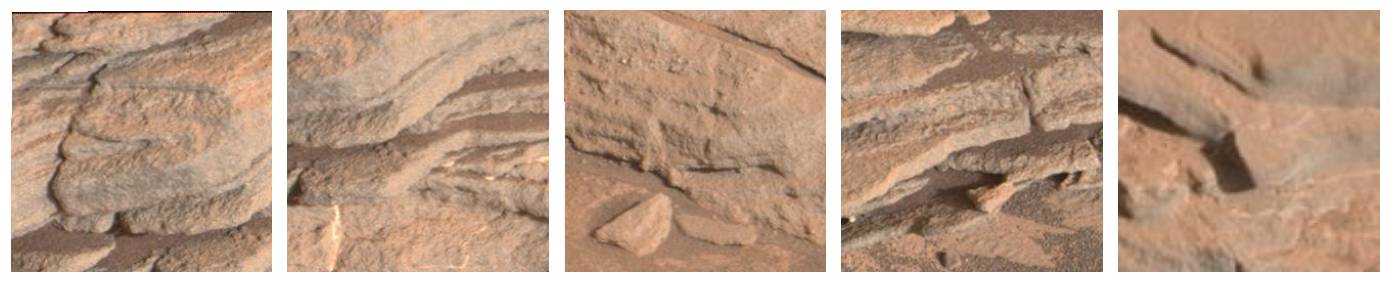}\hspace{0.05\textwidth}
    \includegraphics[width=0.45\textwidth]{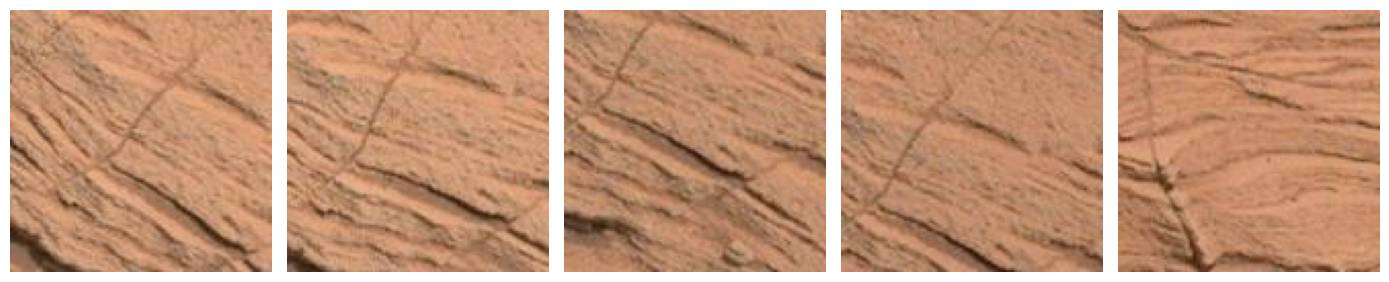}\hspace{0.05\textwidth}
    \includegraphics[width=0.45\textwidth]{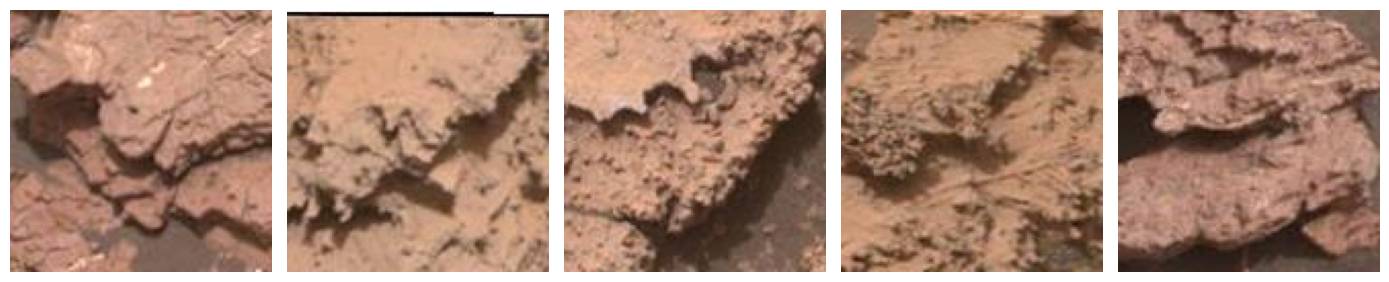}\hspace{0.05\textwidth}
    \includegraphics[width=0.45\textwidth]{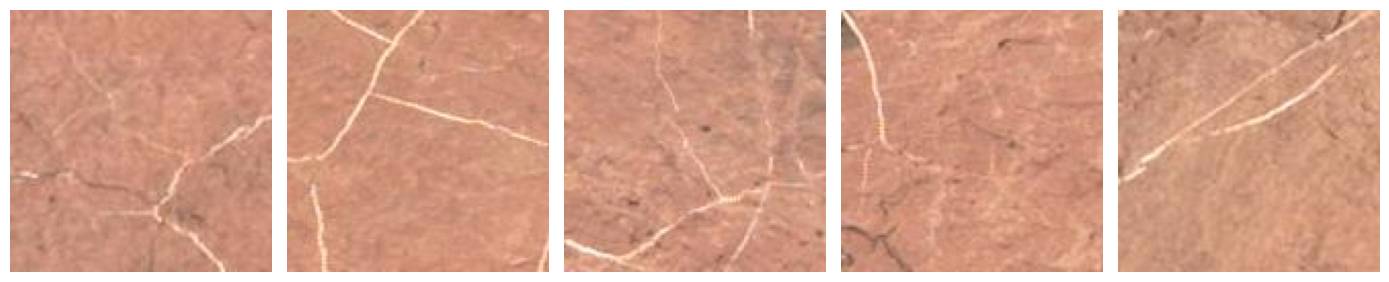}
    \caption{Diverse Martian terrain patches found using deep constrained clustering. Each row consists of two Martian terrains, with five patches from the same cluster arranged in each column. The terrains exhibit a wide range of textures, colors, and geological features showcasing the complexity and diversity of the Martian surface.}
    \label{fig:datasetcluster}
\end{figure*}

\section{Experiments}
\label{ss:exp}
\subsection{Implementation details}
\label{subsec:impl}

We implement our method using either a Deep Encoding Pooling Network (DEP) with an ImageNet-pretrained 18-layer ResNet \cite{resnet} as the CNN backbone or a Vision Transformer (ViT) \cite{vit} for the transformer-based approach, as described in \cref{ss:method}. The dimensionality of the embedding layer is set to $512$. However, DEP is used as the default architecture for most ablation experiments to ensure comparability with prior work \cite{tejasearth}.

For clustering, extracted embeddings are first reduced to 256 dimensions using Principal Component Analysis (PCA), followed by whitening and $\ell^2$ normalization. We employ the PCCClustering algorithm \cite{PCCC}, with the number of clusters ($K$) set to 150 by default unless otherwise specified.

%We also use Faiss K-means clustering algorithm \cite{faiss}, which has been used extensively in prior studies \cite{tejasearth, tejas1}. 

To train the feature extraction networks, we use the SGD optimizer with a learning rate of $1\mathrm{e}{-4}$ (unless specified otherwise) and a weight decay of $1\mathrm{e}{-5}$. The size of a minibatch corresponds to the number of samples per cluster times the number of clusters. We set number of samples per cluster as $4$. Convergence is determined based on cluster stability across epochs, measured using Normalized Mutual Information (NMI) between the cluster assignments at epoch $t$ and $t-1$, as described in \cite{deepcluster}.

Further, all input image patches are resized to a standardized resolution of $224\times224$. As discussed earlier, we do not employ data augmentation techniques during training to preserve the intrinsic geological characteristics of Martian terrain. This consistent experimental framework supports detailed evaluations of cluster quality, including analysis of the number of clusters, architectural differences, and constraints contributions, as discussed in the results section.

%Augmentations such as cropping or color distortions could alter the visual properties of rocks and soil, potentially resulting in unrealistic or scientifically irrelevant representations.

\section{Results}

\label{ss:res}

This section evaluates the performance of the proposed Deep Constrained Clustering with Metric Learning (DCCML) algorithm on the Martian terrain recognition task. We present both quantitative and qualitative clustering evaluations, analyze retrieval performance, and assess the contributions of individual constraint components. Furthermore, we demonstrate the effectiveness of DCCML across different neural network architectures, highlighting its robustness and adaptability.

Our evaluations are motivated by two primary goals: (1) ensuring the formation of clusters that are both compact and scientifically meaningful, and (2) demonstrating that DCCML effectively addresses limitations of existing approaches, such as sensitivity to illumination, texture variations, and color inconsistencies.

\subsection{Clustering Evaluation}

\subsubsection{Quantitative}

The clustering quality is quantitatively evaluated using the Davies-Bouldin Index (DB Index), a metric that assesses the relationship between intra-cluster dispersion and inter-cluster separation. Specifically, the DB Index compares the average spread (dispersion) of points within each cluster to the distance between cluster centroids. A lower DB Index indicates that clusters are compact (low intra-cluster dispersion) and well-separated (high inter-cluster distance), which reflects better clustering performance. 
It is defined as follows:
\[
\text{DB Index} = \frac{1}{n} \sum_{i=1}^{n} \max_{j \neq i} \left( \frac{\sigma_i + \sigma_j}{d(c_i, c_j)} \right),
\]
where $n$ is the number of clusters, $\sigma_i$ and $\sigma_j$ are the average intra-cluster dispersion of $cluster_i$ and $cluster_j$ respectively, and $d(c_i, c_j)$ is the distance between centroids of $cluster_i$ ($c_i$) and $cluster_j$($c_j$).

\subsubsection{Qualitative}
We evaluate the homogeneity of the clusters by randomly visualizing $300$ samples from each cluster. As discussed earlier, we classify a cluster as either uniform or mixed based on whether more than $80\%$ of the patches belong to the same category, as per the taxonomy outlined in \cite{tejasearth}. This analysis highlights the model's ability to group similar terrain features despite variations in texture, illumination, and scale.

\subsection{Retrieval Evaluation}
The performance of the model is further evaluated using a retrieval task, with Precision@$K$ serving as the metric to assess the quality of the retrieved images, similar to the approach described in \cite{tejasearth}. Given a query image $\mathbf I$ with known category outlined in \cite{tejasearth}, we poll the model to retrieve the top-$K$ images $\mathbf{I}_k$ from the test set whose embeddings have the least Euclidean distance to the embedding of the query image. Precision@$K$ is then defined as follows:
\begin{equation}
    \text{Precision@}K \triangleq \frac{1}{K}\sum_{k \in {1..K}} \mathbb{I}(c_{\mathbf{I}} = c_{\mathbf{I_k'}})
\end{equation}
where $\mathbb{I}$ is the indicator function,  $c_{\mathbf I}$ and $c_{\mathbf I_k}$ 

For our evaluation, $K=10$ is used. A high Precision@$K$ indicates that the embeddings effectively map semantically similar terrain patches close together in the embedding space.

\subsection{Performance on Curiosity Dataset}
\begin{figure*}[htb]
\centering
 \begin{minipage}{\textwidth}
    \includegraphics[width=\textwidth, height=8.5cm]{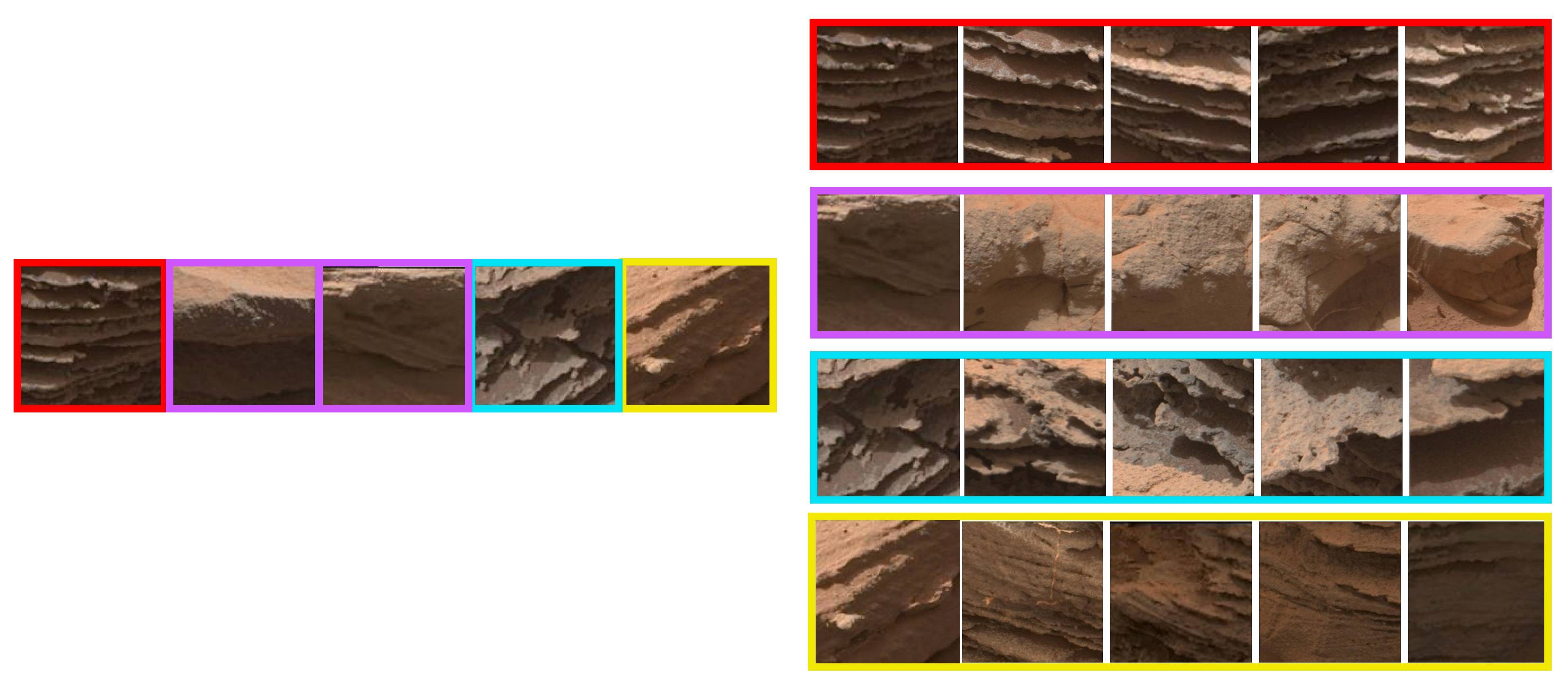}
\end{minipage}
\caption{The left subfigure exhibits a mixed terrain cluster resulting from previous methodology, while the right subfigure illustrates four distinct homogeneous clusters as identified by our improved model, emphasizing terrain classes over attributes such as illumination. }
%The row highlighted by the yellow border demonstrates the model's robustness in handling diverse illumination conditions, ensuring that clustering reflects genuine terrain classes rather than variabilities in lighting.
\label{fig:cusres1}
\end{figure*}
\begin{figure*}[htb]
\centering
 \begin{minipage}{\textwidth}
    \includegraphics[width=\textwidth, height=6.5cm]{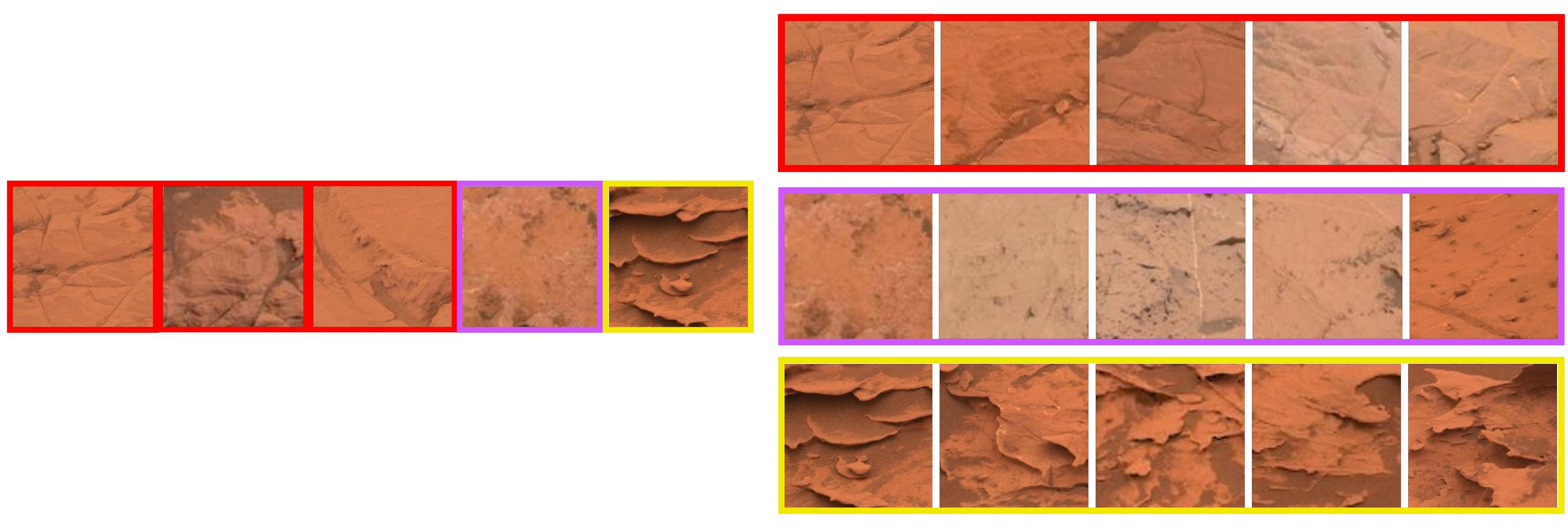}
\end{minipage}
\caption{The left subfigure exhibits a mixed terrain cluster resulting from previous methodology, while the right subfigure illustrates three distinct homogeneous clusters as identified by our improved model, emphasizing terrain classes over attributes such as hue. }
%The row highlighted by the yellow border demonstrates the model's robustness in handling diverse illumination conditions, ensuring that clustering reflects genuine terrain classes rather than variabilities in lighting.
\label{fig:cusres2}
\end{figure*}
\begin{figure*}[htb]
  \centering
  \begin{tabular}{cccc}
    a. & 
    \begin{subfigure}[b]{0.43\textwidth}
    
      \centering
      \includegraphics[width=\textwidth, height=2cm]{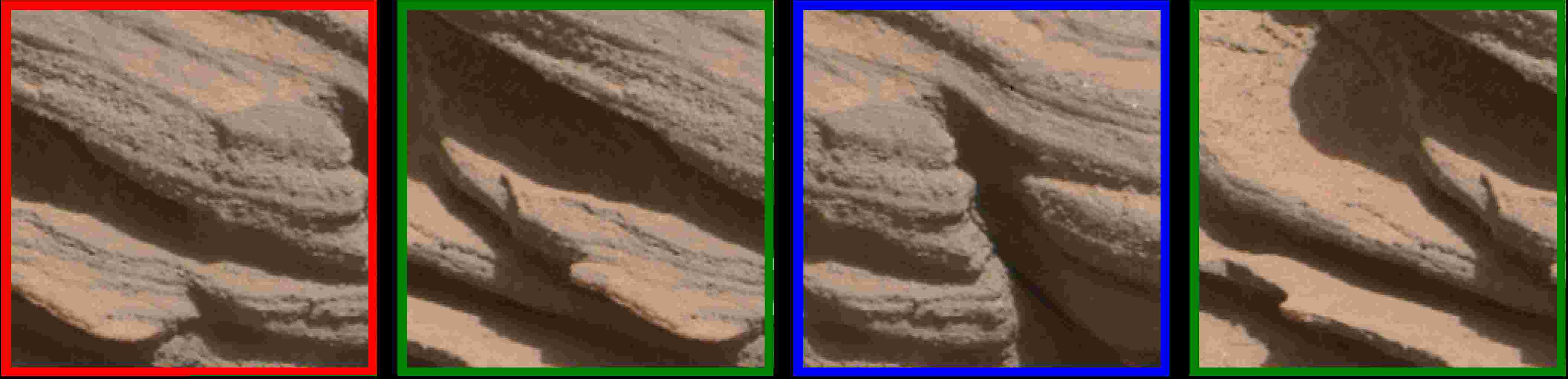}
    \end{subfigure} & b. &
    \begin{subfigure}[b]{0.43\textwidth}
      \centering
      \includegraphics[width=\textwidth, height=2cm]{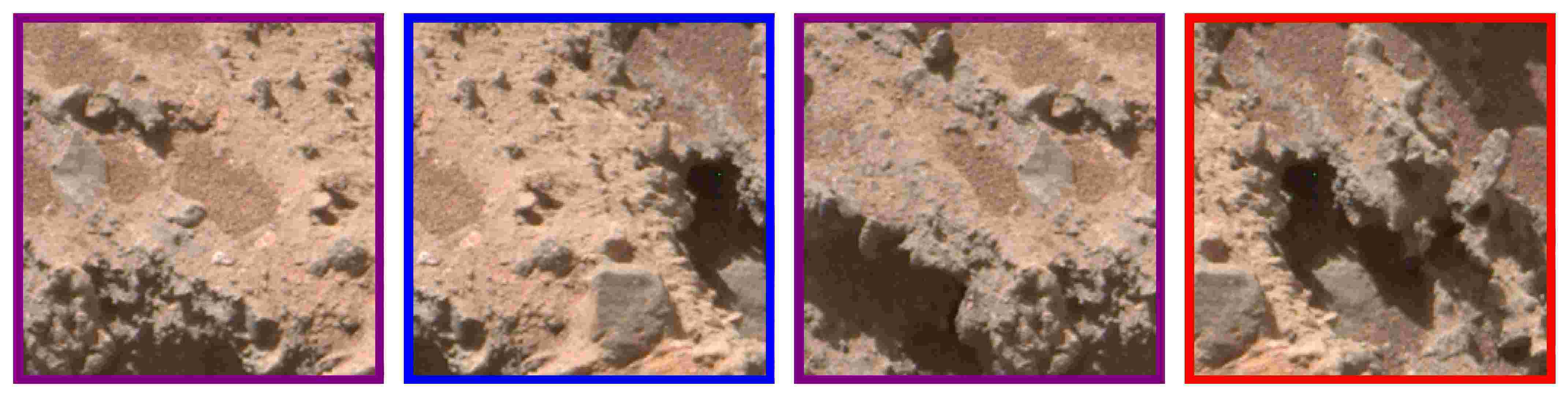}
    \end{subfigure} \\
    c. &
    \begin{subfigure}[b]{0.43\textwidth}
      \centering
      \includegraphics[width=\textwidth, height=2cm]{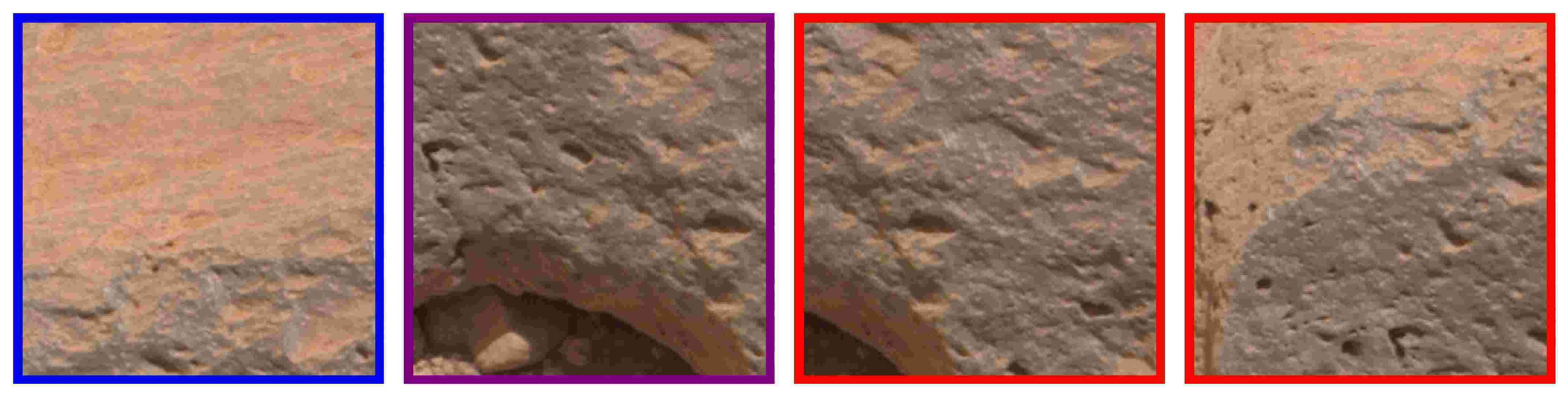}
    \end{subfigure} & d.&
    \begin{subfigure}[b]{0.43\textwidth}
      \centering
      \includegraphics[width=\textwidth, height=2cm]{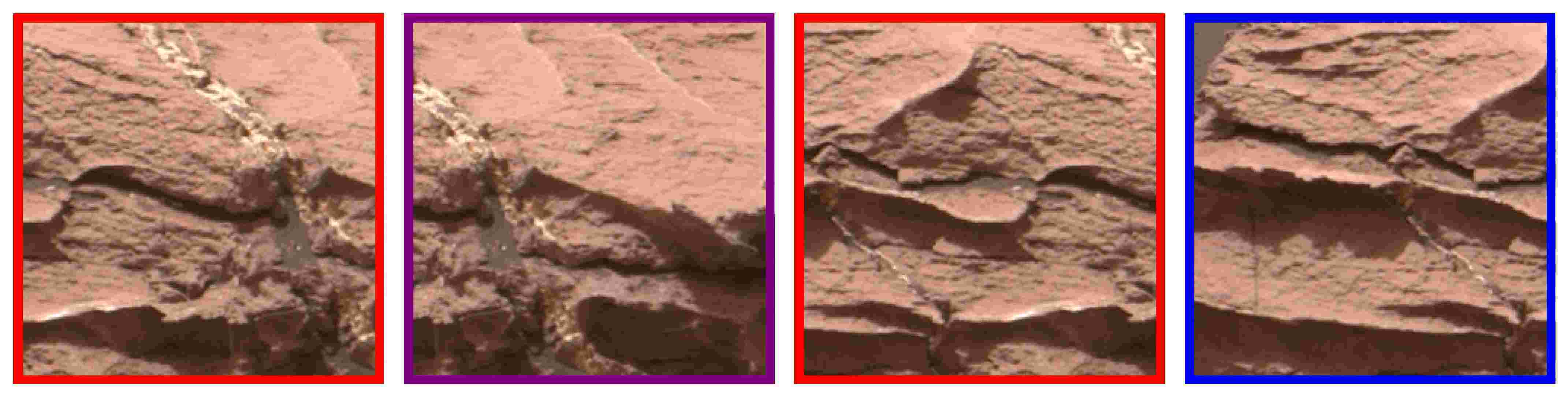}
    \end{subfigure}
  \end{tabular}
  \caption{Four sub-images showcase distinct patches extracted from the same image. Each color border denotes separate clusters as observed in the previous clustering algorithm. However, the current algorithm successfully groups them into a single cluster.}
  \label{figres}
\end{figure*}

 Table \ref{tab:kcc} summarizes the performance comparison of clustering methods on the Curiosity dataset across varying numbers of clusters ($K = 50, 100, 150$). The results demonstrate that our proposed DCCML with all constraints significantly outperforms both DeepCluster \cite{deepcluster} and our previous state-of-the-art, DCML \cite{tejasearth}, in terms of the Davies-Bouldin (DB) Index and the number of homogeneous clusters formed. In particular, at $K = 150$, DCCML achieves a DB Index of $1.82$, reflecting tighter and better-separated clusters compared to DCML’s DB Index of $3.86$. Furthermore, DCCML forms $130$ homogeneous clusters that align closely with the fixed taxonomy, compared to only $105$ homogeneous clusters formed by DCML.

Unlike DCCML, which integrates metric learning and constraint-based clustering, both DeepCluster and DCML rely on Faiss K-means, a clustering method that does not incorporate such constraints. Additionally, DeepCluster does not leverage metric learning and instead relies solely on clustering objectives for feature refinement \cite{deepcluster}. These differences directly impact clustering performance, as seen in Table \ref{tab:kcc}, where DCCML’s constraint-guided clustering leads to tighter, more homogeneous clusters with a significantly lower DB Index. By enforcing spatial, stereo, and RSM constraints, DCCML ensures that clustering decisions are driven by geological features rather than visual artifacts, resulting in semantically meaningful clusters that better align with the predefined taxonomy \cite{tejasearth}.

Beyond clustering performance, retrieval accuracy serves as an additional metric to evaluate the effectiveness of the learned representations. Table \ref{tab:kcc} reports retrieval results alongside clustering metrics, showing that DCCML achieves a retrieval accuracy of $89.86\%$ at $K=150$, significantly surpassing DCML ($86.71\%$) and DeepCluster ($79.52\%$). This improvement is attributed to the integration of pairwise constraints, which guide the clustering process to learn more discriminative embeddings that enhance retrieval. The ability to retrieve geologically similar patches more accurately further highlights the advantage of incorporating domain-specific constraints in clustering.

Also, \cref{fig:datasetcluster} showcases terrain clusters identified by DCCML. Each row represents two taxonomy-defined categories\cite{tejasearth}, with five representative patches from the same cluster displayed for each category. This visualization confirms that DCCML organizes terrain patches into clusters that align with the semantic granularity defined by the taxonomy\cite{tejasearth}, validating its effectiveness in capturing meaningful terrain features.

Further, the left subfigure in \cref{fig:cusres1} illustrates a mixed terrain cluster generated by the DCML, the previous state-of-the-art (SOTA) model, where different terrain types were incorrectly grouped due to sensitivity to factors such as illumination and shadows. This issue highlights the limitations of earlier approaches, which often relied on superficial visual features. In contrast, the proposed DCCML model shown in the right subfigure effectively separates terrain types into four homogeneous clusters that accurately reflect terrain classes from the taxonomy \cite{tejasearth}. For example, the yellow-bordered row demonstrates the model's ability to handle diverse lighting conditions, ensuring that clustering is driven by geological features rather than external visual variations.

Similarly, in \cref{fig:cusres2}, the DCML method grouped the terrain patches by hue, leading to scientifically irrelevant groupings. By incorporating pairwise constraints, the DCCML model resolves this issue by clustering patches based on fundamental geological features such as lamination strength and fracture density, as defined by the taxonomy.

Later, in \cref{figres} (a, b, c, d), the DCML method demonstrates a failure mode where similar terrain patches from the same image were incorrectly assigned to separate clusters due to variations in illumination, shadow, and texture. Each sub-image in \cref{figres} contains four similar terrain patches, but the baseline method partitions them into three distinct clusters, as indicated by the color borders. In contrast, the proposed method effectively resolves this issue by grouping all similar patches within each sub-image into a single cohesive cluster.

\begin{table}[H]
    \centering
    \caption{Performance of clustering methods with varying number of clusters on Curiosity dataset}
    \begin{tabularx}{\linewidth}{lXXXX}
        \toprule
        Algorithm & No. of  & DB Index & No. of & Retrieval \\
        &Clusters&&Homo-&Results\\&&&geneous Clusters&\\
        
        \midrule
        DeepCluster \cite{deepcluster} 
         & 50 & 5.3 & 16 & 61.43 \\
         & 100 & 4.98 & 55 & 70.95 \\
         & 150 & 4.36 & 94 & 79.52\\
        \midrule
        DCML \cite{tejasearth} 
         & 50 & 5.2 & 21 & 64.76\\
         & 100 & 3.9 & 68 & 74.28\\
         & 150 & 3.86 & 105 & 86.71\\
        \midrule
        DCCML with
         & 50 & 2.72 & 39 & 70.48\\
         All Constraints & 100 & 2.16 & 82 & 81.90 \\
         (Our method) & \textbf{150} & \textbf{1.82} & \textbf{130} & \textbf{89.86} \\
        \bottomrule
    \end{tabularx}
    \label{tab:kcc}
\end{table}

\subsection{Analyzing Constraint Contributions}

Table \ref{tab:Constraints} analyzes the contribution of each constraint component individually in the DCCML algorithm. From Table \ref{tab:Constraints}, it is observed that incorporating all constraints (Neighboring, LR, and RSM) into the DCCML method achieves the best performance, with a DB Index of $1.82$ with $130$ homogeneous clusters, and a $89.86\%$ retrieval accuracy. By linking spatially close patches with similar depths, neighboring constraints ensure that rock formations with similar features remain grouped together, avoiding fragmentation caused by minor variations such as shadow or dust differences. Meanwhile, LR constraints bring patches from stereo images covering the same area closer together, counteracting scale and sharpness mismatches and making the embedded space scale invariant. Although RSM constraints show smaller individual gains, they link patches across consecutive images, mitigating the impact of minor viewpoint or illumination differences and preventing clusters from splitting unnecessarily.

The combination of Neighboring and LR constraints also performs competitively, with a DB Index of 2.08, 130 homogeneous clusters, and a retrieval accuracy of 89.64\%. This highlights the spatial consistency (Neighboring constraints) and resolution invariance (LR constraints) is the most crucial in improving clustering quality. Overall, the joint effect of these constraints addresses challenges like shadow variations, resolution mismatches, and temporal changes, resulting in superior cluster separation, increased homogeneity, and improved retrieval accuracy compared to DCML and partially constrained variants.
\begin{table}[H]
    \centering
    \caption{Analyzing the contribution of each constraint component individually}
    \begin{tabularx}{\linewidth}{l>{\raggedright\arraybackslash}X>{\raggedright\arraybackslash}X>{\raggedright\arraybackslash}X}
        \toprule
        Method & DB Index & No. of Homogeneous Clusters & Retrieval Results\\
        \midrule
        DCML & 3.86 & 105 & 86.71\\
        DCCML+Neighbouring Constraints & 1.97 & 121 & 89.04\\
        DCCML+LR Constraints & 2.24 & 114 & 87.82\\
        DCCML+RSM Constraints & 3.4 & 108 & 87.37\\
        DCCML+Neighb.+LR Constr. & 2.08 & 130 & 89.64\\
        \textbf{DCCML+All Constraints} & \textbf{1.82} & \textbf{130} & \textbf{89.86}\\
        \bottomrule
    \end{tabularx}
    \label{tab:Constraints}
\end{table}

\subsection{Cluster Evaluation Results with Different Architecture}

\begin{table}[H]
    \centering
    \caption{Cluster Evaluation with different architecture}
    \begin{tabularx}{\linewidth}{XXXX}
        \toprule
        Method & DB Index & Num of & Retrieval\\
               &          & Homogeneous & Results\\
               &          & Clusters & \\
        \midrule
        DEP  & 1.82 & $\sim$130 & 89.86\\
        ViT & \textbf{1.74} & $\sim$130 & \textbf{90.47}\\
        \bottomrule
    \end{tabularx}
    \label{tab:clustering_results}
\end{table}

While our primary experiments use the Deep Encoding Pooling (DEP) architecture \cite{gtos-dep} for continuity with prior Martian terrain studies, we also compare it against the Vision Transformer (ViT) \cite{vit}, a more recent state-of-the-art paradigm for various computer vision tasks. \cref{tab:clustering_results} shows that both backbones yield comparable clustering results at $K=150$, with ViT achieving a slightly lower DB Index of $1.74$ compared to $1.82$ for DEP, and each forming approximately 130 homogeneous clusters. This consistency across architectures highlights DCCML’s flexibility for Martian terrain recognition and its ability to generate meaningful, cohesive clusters as long as the feature extraction model provides sufficiently rich representations. Moreover, retrieval performance follows a similar trend, with ViT achieving a retrieval accuracy of $90.47\%$ compared to $89.86\%$ for DEP. While ViT shows a slight advantage, the overall consistency across architectures reinforces DCCML’s robustness in clustering and retrieval tasks, regardless of the feature extraction model used.

\section{Conclusion}
\label{ss:conclusion}

This paper presents Deep Constrained Clustering with Metric Learning (DCCML), a novel algorithm for Martian terrain recognition that effectively discovers geologically meaningful clusters. By integrating pairwise constraints derived from spatial proximity, stereo image correspondences, and RSM image pairs, DCCML mitigates the challenges posed by variations in intensity, scale, rotation, and illumination. Extensive evaluations on the Curiosity rover dataset demonstrate that DCCML outperforms prior deep clustering methods, achieving higher cluster homogeneity and improved terrain classification. These findings, validated through quantitative DB Index metrics and qualitative visual inspection, confirm DCCML’s ability to identify distinct geological units within Martian imagery.

Unlike previous methods that struggled with illumination differences, resolution mismatches, and depth ambiguities, DCCML produces clusters that align with intrinsic geological structures, capturing meaningful textures and morphological characteristics rather than superficial visual artifacts. This capability is crucial for planetary science, as it enables an automated and scalable approach for categorizing terrain types and analyzing Martian geology.

Beyond scientific analysis, DCCML has significant implications for autonomous planetary exploration. The geological clusters it generates can be used to identify high-value sampling targets, optimize rover navigation strategies, and refine site selection for in-situ experiments. By enabling unsupervised terrain classification, DCCML supports autonomous decision-making processes that maximize scientific return in future Mars missions.

To further enhance its applicability, DCCML will be extended to analyze terrain data from additional Martian missions, such as Perseverance, to evaluate its robustness across diverse geological settings. Additionally, expert geologists will label the clusters generated from Curiosity data, creating a high-quality labeled dataset that will serve as a foundation for Generalized Class Discovery (GCD). This extension will enable the identification of unknown terrain categories, expanding our ability to analyze unlabeled Martian imagery in future missions.

Ultimately, DCCML establishes a scalable and adaptable clustering framework for planetary science, advancing our understanding of Martian geology while laying the groundwork for enhanced autonomous exploration and geological discovery.

% or


%\section*{Acknowledgment}


\ifCLASSOPTIONcaptionsoff
  \newpage
\fi
\newpage

\begin{IEEEbiography}
[{\includegraphics[width=1in,height=1.25in,clip,keepaspectratio]{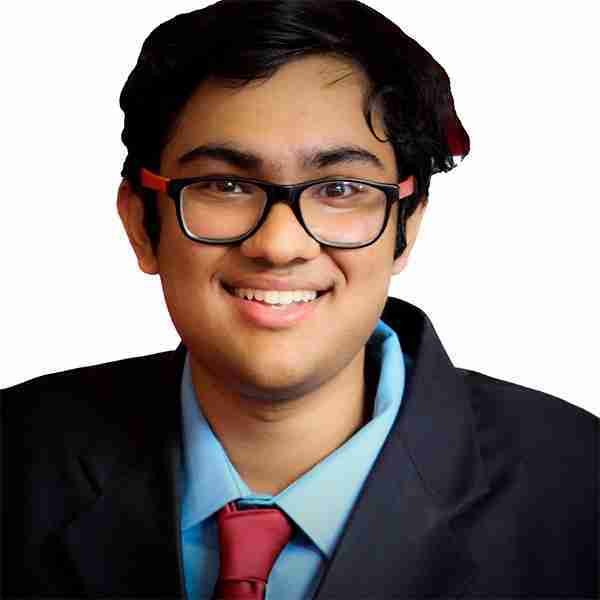}}]{Tejas Panambur} is a Ph.D. candidate in Electrical and Computer Engineering at the University of Massachusetts, Amherst, with over 5 years of experience in machine learning, deep learning, and computer vision. As a Research Assistant in the Remote Hyperspectral Observers (RHO) group, his work focuses on self-supervised learning and deep clustering to improve the analysis of Mars mission imagery. Tejas has contributed to significant advancements in Martian terrain categorization and planetary exploration. He has also conducted research internships in deep active learning at Nokia Bell Labs and worked on foundational deep learning models for lunar mapping at the Search for Extraterrestrial Intelligence (SETI). His expertise spans image processing, deep learning, and planetary data analysis.
\end{IEEEbiography}

\begin{IEEEbiography}
[{\includegraphics[width=1in,height=1.25in,clip,keepaspectratio]{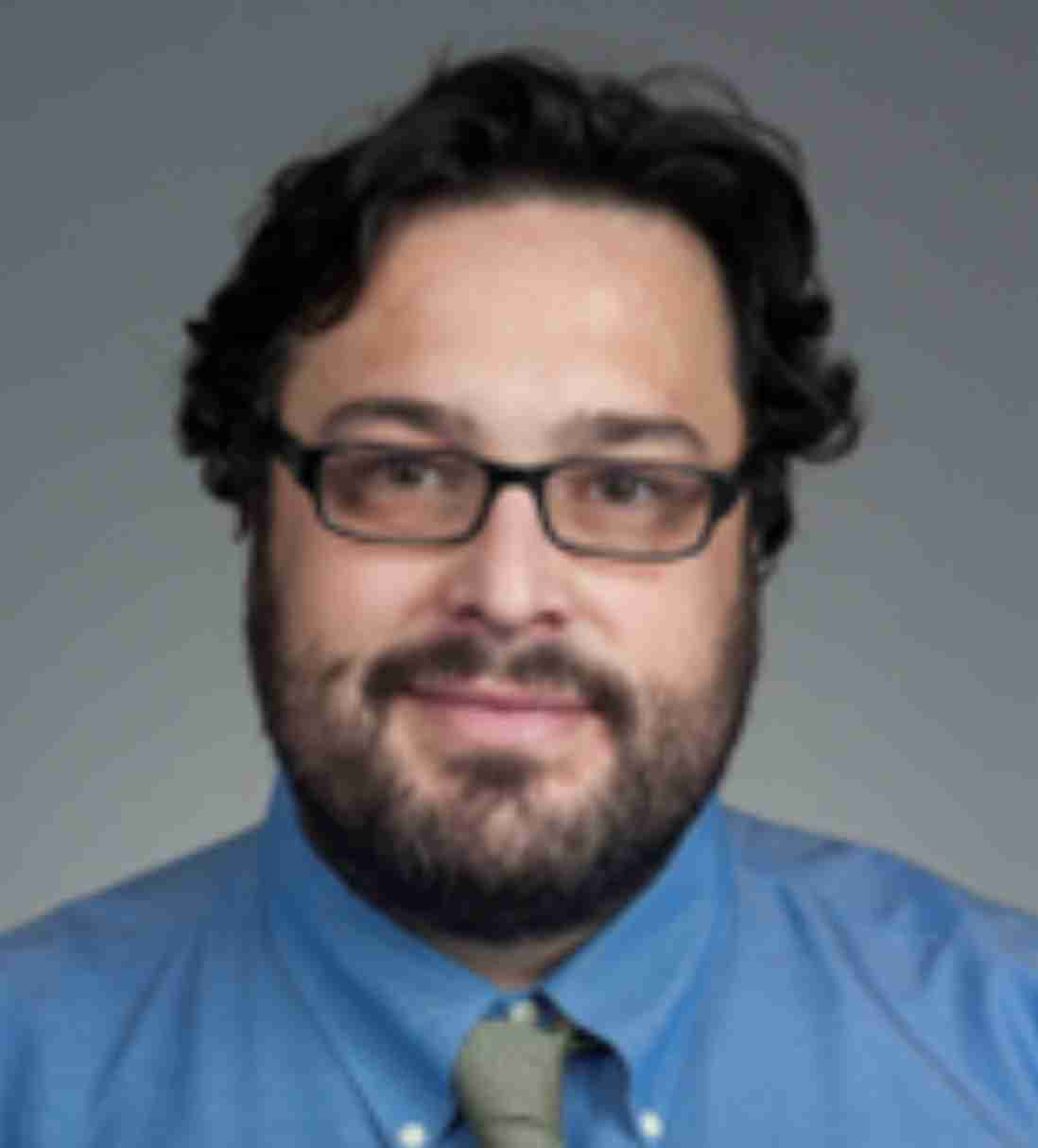}}]{Mario Parente} (M’05–SM’13) received the B.S. and M.S. (summa cum laude) degrees in telecommunication engineering from the University of Naples Federico II, Naples, Italy, in 2001 and 2003, respectively, and the M.S. and Ph.D. degrees in electrical engineering from Stanford University, Stanford, CA, USA, in 2005 and 2010, respectively.

He was a Postdoctoral Associate in the Department of Geosciences, Brown University, Providence, RI, USA. He is currently an Associate Professor with the Department of Electrical and Computer Engineering, University of Massachusetts, Amherst, MA, USA. He has supported several scientific teams in NASA missions such as the Compact Reconnaissance Imaging Spectrometer for Mars, the Mars Mineralogy Mapper, and the Mars Science Laboratory ChemCam science teams. He is a Principal Investigator at the SETI Institute, Carl Sagan Center for Search for Life in the Universe and a member of the NASA Astrobiology Institute. His research involves combining physical models and statistical techniques to address issues in remote sensing of Earth and planetary surfaces. His professional interests include identification of ground composition, geomorphological feature detection and imaging spectrometer data modeling, and reduction and calibration for NASA missions. He developed machine learning algorithms for the representation and processing of hyperspectral data based on statistical, geometrical, and topological models. His research also involves the study of physical models of light scattering in particulate media. Furthermore, he has developed solutions for the integration of color and hyperspectral imaging and robotics to identify scientifically significant targets for rover and orbiter-based reconnaissance. Dr. Parente is an Associate Editor for the IEEE Geoscience and Remote Sensing Magazine and the IEEE Geoscience and Remote Sensing Letters.

\end{IEEEbiography}

%\newpage


%\vfill



\begin{thebibliography}{10}
\providecommand{\url}[1]{#1}
\csname url@samestyle\endcsname
\providecommand{\newblock}{\relax}
\providecommand{\bibinfo}[2]{#2}
\providecommand{\BIBentrySTDinterwordspacing}{\spaceskip=0pt\relax}
\providecommand{\BIBentryALTinterwordstretchfactor}{4}
\providecommand{\BIBentryALTinterwordspacing}{\spaceskip=\fontdimen2\font plus
\BIBentryALTinterwordstretchfactor\fontdimen3\font minus \fontdimen4\font\relax}
\providecommand{\BIBforeignlanguage}[2]{{%
\expandafter\ifx\csname l@#1\endcsname\relax
\typeout{** WARNING: IEEEtran.bst: No hyphenation pattern has been}%
\typeout{** loaded for the language `#1'. Using the pattern for}%
\typeout{** the default language instead.}%
\else
\language=\csname l@#1\endcsname
\fi
#2}}
\providecommand{\BIBdecl}{\relax}
\BIBdecl

\bibitem{msl}
J.~P. Grotzinger, J.~Crisp, A.~R. Vasavada, R.~C. Anderson, C.~J. Baker, R.~Barry, D.~F. Blake, P.~Conrad, K.~S. Edgett, B.~Ferdowski \emph{et~al.}, ``Mars science laboratory mission and science investigation,'' \emph{Space science reviews}, vol. 170, no.~1, pp. 5--56, 2012.

\bibitem{mars2020}
K.~H. Williford, K.~A. Farley, K.~M. Stack, A.~C. Allwood, D.~Beaty, L.~W. Beegle, R.~Bhartia, A.~J. Brown, M.~de~la Torre~Juarez, S.-E. Hamran \emph{et~al.}, ``The nasa mars 2020 rover mission and the search for extraterrestrial life,'' in \emph{From habitability to life on Mars}.\hskip 1em plus 0.5em minus 0.4em\relax Elsevier, 2018, pp. 275--308.

\bibitem{Bell2017-ia}
J.~F. Bell, III et~al., ``The mars science laboratory curiosity rover mastcam instruments: Preflight and in-flight calibration, validation, and data archiving : {MSL/Mastcam} calibration,'' \emph{Earth and Space Science}, vol.~4, no.~7, pp. 396--452, Jul. 2017.

\bibitem{Malin2017-vg}
M.~C. Malin~et. al, ``The mars science laboratory ({MSL}) mast cameras and descent imager: Investigation and instrument descriptions,'' \emph{Earth and Space Science}, vol.~4, no.~8, pp. 506--539, Aug. 2017.

\bibitem{scoti}
\BIBentryALTinterwordspacing
D.~Qiu, B.~Rothrock, T.~Islam, A.~K. Didier, V.~Z. Sun, C.~A. Mattmann, and M.~Ono, ``Scoti: Science captioning of terrain images for data prioritization and local image search,'' \emph{Planetary and Space Science}, vol. 188, p. 104943, 2020. [Online]. Available: \url{https://www.sciencedirect.com/science/article/pii/S0032063319301242}
\BIBentrySTDinterwordspacing

\bibitem{chakravarthy2021mrscatt}
A.~S. Chakravarthy, R.~Roy, and P.~Ravirathinam, ``Mrscatt: A spatio-channel attention-guided network for mars rover image classification,'' in \emph{Proceedings of the IEEE/CVF Conference on Computer Vision and Pattern Recognition}, 2021, pp. 1961--1970.

\bibitem{mslv1}
K.~Wagstaff, Y.~Lu, A.~Stanboli, K.~Grimes, T.~Gowda, and J.~Padams, ``Deep mars: Cnn classification of mars imagery for the pds imaging atlas,'' in \emph{Proceedings of the AAAI Conference on Artificial Intelligence}, vol.~32, no.~1, 2018.

\bibitem{mslv2}
K.~Wagstaff, S.~Lu, E.~Dunkel, K.~Grimes, B.~Zhao, J.~Cai, S.~B. Cole, G.~Doran, R.~Francis, J.~Lee \emph{et~al.}, ``Mars image content classification: Three years of nasa deployment and recent advances,'' in \emph{Proceedings of the AAAI Conference on Artificial Intelligence}, vol.~35, no.~17, 2021, pp. 15\,204--15\,213.

\bibitem{wang2021semi}
W.~Wang, L.~Lin, Z.~Fan, and J.~Liu, ``Semi-supervised learning for mars imagery classification,'' in \emph{2021 IEEE International Conference on Image Processing (ICIP)}.\hskip 1em plus 0.5em minus 0.4em\relax IEEE, 2021, pp. 499--503.

\bibitem{gonzalez2018deepterramechanics}
R.~Gonzalez and K.~Iagnemma, ``Deepterramechanics: Terrain classification and slip estimation for ground robots via deep learning,'' \emph{arXiv preprint arXiv:1806.07379}, 2018.

\bibitem{PALAFOX201748}
\BIBentryALTinterwordspacing
L.~F. Palafox, C.~W. Hamilton, S.~P. Scheidt, and A.~M. Alvarez, ``Automated detection of geological landforms on mars using convolutional neural networks,'' \emph{Computers \& Geosciences}, vol. 101, pp. 48--56, 2017. [Online]. Available: \url{https://www.sciencedirect.com/science/article/pii/S0098300416305532}
\BIBentrySTDinterwordspacing

\bibitem{spoc}
B.~Rothrock, R.~Kennedy, C.~Cunningham, J.~Papon, M.~Heverly, and M.~Ono, ``Spoc: Deep learning-based terrain classification for mars rover missions,'' in \emph{AIAA SPACE 2016}, 2016, p. 5539.

\bibitem{ai4mars}
R.~M. Swan, D.~Atha, H.~A. Leopold, M.~Gildner, S.~Oij, C.~Chiu, and M.~Ono, ``Ai4mars: A dataset for terrain-aware autonomous driving on mars,'' in \emph{Proceedings of the IEEE/CVF Conference on Computer Vision and Pattern Recognition}, 2021, pp. 1982--1991.

\bibitem{jplcontrastive}
I.~R. Ward, C.~Moore, K.~Pak, J.~Chen, and E.~Goh, ``Improving contrastive learning on visually homogeneous mars rover images,'' 2022.

\bibitem{tejas1}
M.~Parente and T.~Panambur, ``Classification of martian terrains via deep clustering of mastcam images,'' in \emph{IGARSS 2020-2020 IEEE International Geoscience and Remote Sensing Symposium}.\hskip 1em plus 0.5em minus 0.4em\relax IEEE, 2020, pp. 1054--1057.

\bibitem{tejas2}
T.~Panambur and M.~Parente, ``Improved deep clustering of mastcam images using metric learning,'' in \emph{2021 IEEE International Geoscience and Remote Sensing Symposium IGARSS}.\hskip 1em plus 0.5em minus 0.4em\relax IEEE, 2021, pp. 2859--2862.

\bibitem{tejasearth}
T.~Panambur, D.~Chakraborty, M.~Meyer, R.~Milliken, E.~Learned-Miller, and M.~Parente, ``Self-supervised learning to guide scientifically relevant categorization of martian terrain images,'' in \emph{Proceedings of the IEEE/CVF Conference on Computer Vision and Pattern Recognition (CVPR) Workshops}, June 2022, pp. 1322--1332.

\bibitem{simclr}
T.~Chen, S.~Kornblith, M.~Norouzi, and G.~Hinton, ``A simple framework for contrastive learning of visual representations,'' in \emph{International conference on machine learning}.\hskip 1em plus 0.5em minus 0.4em\relax PMLR, 2020, pp. 1597--1607.

\bibitem{moco}
K.~He, H.~Fan, Y.~Wu, S.~Xie, and R.~Girshick, ``Momentum contrast for unsupervised visual representation learning,'' in \emph{Proceedings of the IEEE/CVF conference on computer vision and pattern recognition}, 2020, pp. 9729--9738.

\bibitem{swav}
M.~Caron, I.~Misra, J.~Mairal, P.~Goyal, P.~Bojanowski, and A.~Joulin, ``Unsupervised learning of visual features by contrasting cluster assignments,'' \emph{Advances in Neural Information Processing Systems}, vol.~33, pp. 9912--9924, 2020.

\bibitem{byol}
J.-B. Grill, F.~Strub, F.~Altch{\'e}, C.~Tallec, P.~Richemond, E.~Buchatskaya, C.~Doersch, B.~Avila~Pires, Z.~Guo, M.~Gheshlaghi~Azar \emph{et~al.}, ``Bootstrap your own latent-a new approach to self-supervised learning,'' \emph{Advances in Neural Information Processing Systems}, vol.~33, pp. 21\,271--21\,284, 2020.

\bibitem{simsiam}
X.~Chen and K.~He, ``Exploring simple siamese representation learning,'' in \emph{Proceedings of the IEEE/CVF Conference on Computer Vision and Pattern Recognition}, 2021, pp. 15\,750--15\,758.

\bibitem{barlow}
J.~Zbontar, L.~Jing, I.~Misra, Y.~LeCun, and S.~Deny, ``Barlow twins: Self-supervised learning via redundancy reduction,'' in \emph{International Conference on Machine Learning}.\hskip 1em plus 0.5em minus 0.4em\relax PMLR, 2021, pp. 12\,310--12\,320.

\bibitem{colorization}
G.~Larsson, M.~Maire, and G.~Shakhnarovich, ``Colorization as a proxy task for visual understanding,'' in \emph{Proceedings of the IEEE conference on computer vision and pattern recognition}, 2017, pp. 6874--6883.

\bibitem{jigsaw}
M.~Noroozi and P.~Favaro, ``Unsupervised learning of visual representations by solving jigsaw puzzles,'' in \emph{European conference on computer vision}.\hskip 1em plus 0.5em minus 0.4em\relax Springer, 2016, pp. 69--84.

\bibitem{rotation}
N.~Komodakis and S.~Gidaris, ``Unsupervised representation learning by predicting image rotations,'' in \emph{International Conference on Learning Representations (ICLR)}, 2018.

\bibitem{doersch2016}
\BIBentryALTinterwordspacing
C.~Doersch, A.~Gupta, and A.~A. Efros, ``Unsupervised visual representation learning by context prediction,'' 2016. [Online]. Available: \url{https://arxiv.org/abs/1505.05192}
\BIBentrySTDinterwordspacing

\bibitem{dino}
\BIBentryALTinterwordspacing
M.~Caron, H.~Touvron, I.~Misra, H.~Jégou, J.~Mairal, P.~Bojanowski, and A.~Joulin, ``Emerging properties in self-supervised vision transformers,'' 2021. [Online]. Available: \url{https://arxiv.org/abs/2104.14294}
\BIBentrySTDinterwordspacing

\bibitem{deepcluster}
M.~Caron, P.~Bojanowski, A.~Joulin, and M.~Douze, ``Deep clustering for unsupervised learning of visual features,'' in \emph{Proceedings of the European conference on computer vision (ECCV)}, 2018, pp. 132--149.

\bibitem{wagstaffconstrained}
K.~Wagstaff, C.~Cardie, S.~Rogers, S.~Schr{\"o}dl \emph{et~al.}, ``Constrained k-means clustering with background knowledge,'' in \emph{Icml}, vol.~1, 2001, pp. 577--584.

\bibitem{ccml}
M.~Bilenko, S.~Basu, and R.~J. Mooney, ``Integrating constraints and metric learning in semi-supervised clustering,'' in \emph{Proceedings of the twenty-first international conference on Machine learning}.\hskip 1em plus 0.5em minus 0.4em\relax ACM, 2004, p.~11.

\bibitem{ccsa}
\BIBentryALTinterwordspacing
Z.~Lu and M.~A. Carreira-Perpinan, ``Constrained spectral clustering through affinity propagation.'' in \emph{CVPR}.\hskip 1em plus 0.5em minus 0.4em\relax IEEE Computer Society, 2008. [Online]. Available: \url{http://dblp.uni-trier.de/db/conf/cvpr/cvpr2008.html#LuC08}
\BIBentrySTDinterwordspacing

\bibitem{probframework}
S.~Basu, M.~Bilenko, and R.~J. Mooney, ``A probabilistic framework for semi-supervised clustering,'' in \emph{Proceedings of the tenth ACM SIGKDD international conference on Knowledge discovery and data mining}, 2004, pp. 59--68.

\bibitem{XingNJR02}
\BIBentryALTinterwordspacing
E.~P. Xing, A.~Y. Ng, M.~I. Jordan, and S.~J. Russell, ``Distance metric learning with application to clustering with side-information.'' in \emph{NIPS}, S.~Becker, S.~Thrun, and K.~Obermayer, Eds.\hskip 1em plus 0.5em minus 0.4em\relax MIT Press, 2002, pp. 505--512. [Online]. Available: \url{http://dblp.uni-trier.de/db/conf/nips/nips2002.html#XingNJR02}
\BIBentrySTDinterwordspacing

\bibitem{zhang2019}
H.~Zhang, S.~Basu, and I.~Davidson, ``A framework for deep constrained clustering -- algorithms and advances,'' 2019.

\bibitem{hsu2016neural}
Y.-C. Hsu and Z.~Kira, ``Neural network-based clustering using pairwise constraints,'' 2016.

\bibitem{shaham2018}
U.~Shaham, K.~Stanton, H.~Li, B.~Nadler, R.~Basri, and Y.~Kluger, ``Spectralnet: Spectral clustering using deep neural networks,'' 2018.

\bibitem{guo2017improved}
X.~Guo, L.~Gao, X.~Liu, and J.~Yin, ``Improved deep embedded clustering with local structure preservation.'' in \emph{Ijcai}, vol.~17, 2017, pp. 1753--1759.

\bibitem{ren2022deep}
Y.~Ren, J.~Pu, Z.~Yang, J.~Xu, G.~Li, X.~Pu, P.~S. Yu, and L.~He, ``Deep clustering: A comprehensive survey,'' 2022.

\bibitem{dec2016}
\BIBentryALTinterwordspacing
J.~Xie, R.~Girshick, and A.~Farhadi, ``Unsupervised deep embedding for clustering analysis,'' in \emph{Proceedings of The 33rd International Conference on Machine Learning}, ser. Proceedings of Machine Learning Research, M.~F. Balcan and K.~Q. Weinberger, Eds., vol.~48.\hskip 1em plus 0.5em minus 0.4em\relax New York, New York, USA: PMLR, 20--22 Jun 2016, pp. 478--487. [Online]. Available: \url{https://proceedings.mlr.press/v48/xieb16.html}
\BIBentrySTDinterwordspacing

\bibitem{PCCC}
\BIBentryALTinterwordspacing
P.~Baumann and D.~S. Hochbaum, ``Pccc: The pairwise-confidence-constraints-clustering algorithm,'' \emph{ArXiv}, vol. abs/2212.14437, 2022. [Online]. Available: \url{https://api.semanticscholar.org/CorpusID:255340906}
\BIBentrySTDinterwordspacing

\bibitem{labelmars}
S.~Schwenzer, M.~Woods, S.~Karachalios, N.~Phan, and L.~Joudrier, ``Labelmars: Creating an extremely large martian image dataset through machine learning,'' in \emph{50th Annual Lunar and Planetary Science Conference}, no. 2132, 2019, p. 1970.

\bibitem{demud}
K.~L. Wagstaff, N.~L. Lanza, D.~R. Thompson, T.~G. Dietterich, and M.~S. Gilmore, ``Guiding scientific discovery with explanations using demud,'' in \emph{Twenty-Seventh AAAI Conference on Artificial Intelligence}, 2013.

\bibitem{mixedpatch}
T.~Panambur and M.~Parente, ``Improved self-supervised texture recognition of mastcam images by eliminating mixed terrain and range patches,'' in \emph{IGARSS 2023 - 2023 IEEE International Geoscience and Remote Sensing Symposium}, 2023, pp. 7292--7295.

\bibitem{gtos-dep}
J.~Xue, H.~Zhang, and K.~Dana, ``Deep texture manifold for ground terrain recognition,'' in \emph{Proceedings of the IEEE Conference on Computer Vision and Pattern Recognition}, 2018, pp. 558--567.

\bibitem{vit}
\BIBentryALTinterwordspacing
A.~Dosovitskiy, L.~Beyer, A.~Kolesnikov, D.~Weissenborn, X.~Zhai, T.~Unterthiner, M.~Dehghani, M.~Minderer, G.~Heigold, S.~Gelly, J.~Uszkoreit, and N.~Houlsby, ``An image is worth 16x16 words: Transformers for image recognition at scale,'' \emph{CoRR}, vol. abs/2010.11929, 2020. [Online]. Available: \url{https://arxiv.org/abs/2010.11929}
\BIBentrySTDinterwordspacing

\bibitem{depten}
\BIBentryALTinterwordspacing
H.~Zhang, J.~Xue, and K.~J. Dana, ``Deep {TEN:} texture encoding network,'' \emph{CoRR}, vol. abs/1612.02844, 2016. [Online]. Available: \url{http://arxiv.org/abs/1612.02844}
\BIBentrySTDinterwordspacing

\bibitem{bilinear}
\BIBentryALTinterwordspacing
T.~Lin and S.~Maji, ``Improved bilinear pooling with cnns,'' \emph{CoRR}, vol. abs/1707.06772, 2017. [Online]. Available: \url{http://arxiv.org/abs/1707.06772}
\BIBentrySTDinterwordspacing

\bibitem{midas}
R.~Ranftl, K.~Lasinger, D.~Hafner, K.~Schindler, and V.~Koltun, ``Towards robust monocular depth estimation: Mixing datasets for zero-shot cross-dataset transfer,'' 2020.

\bibitem{Lowe}
\BIBentryALTinterwordspacing
D.~G. Lowe, ``Distinctive image features from scale-invariant keypoints,'' \emph{Int. J. Comput. Vision}, vol.~60, no.~2, pp. 91--110, Nov. 2004. [Online]. Available: \url{http://dx.doi.org/10.1023/B:VISI.0000029664.99615.94}
\BIBentrySTDinterwordspacing

\bibitem{Hartley2004}
R.~I. Hartley and A.~Zisserman, \emph{Multiple View Geometry in Computer Vision}, 2nd~ed.\hskip 1em plus 0.5em minus 0.4em\relax Cambridge University Press, ISBN: 0521540518, 2004.

\bibitem{malin2017mars}
M.~C. Malin, M.~A. Ravine, M.~A. Caplinger, F.~Tony~Ghaemi, J.~A. Schaffner, J.~N. Maki, J.~F. Bell~III, J.~F. Cameron, W.~E. Dietrich, K.~S. Edgett \emph{et~al.}, ``The mars science laboratory (msl) mast cameras and descent imager: Investigation and instrument descriptions,'' \emph{Earth and Space Science}, vol.~4, no.~8, pp. 506--539, 2017.

\bibitem{grounding-dino}
S.~Liu, Z.~Zeng, T.~Ren, F.~Li, H.~Zhang, J.~Yang, C.~Li, J.~Yang, H.~Su, J.~Zhu \emph{et~al.}, ``Grounding dino: Marrying dino with grounded pre-training for open-set object detection,'' \emph{arXiv preprint arXiv:2303.05499}, 2023.

\bibitem{SAM}
A.~Kirillov, E.~Mintun, N.~Ravi, H.~Mao, C.~Rolland, L.~Gustafson, T.~Xiao, S.~Whitehead, A.~C. Berg, W.-Y. Lo, P.~Doll{\'a}r, and R.~Girshick, ``Segment anything,'' \emph{arXiv:2304.02643}, 2023.

\bibitem{resnet}
K.~He, X.~Zhang, S.~Ren, and J.~Sun, ``Deep residual learning for image recognition,'' in \emph{Proceedings of the IEEE conference on computer vision and pattern recognition}, 2016, pp. 770--778.

\end{thebibliography}
\end{document}